%% file: iclr2026_conference.tex
\documentclass{article} % For LaTeX2e
\usepackage{iclr2026_conference,times}

\input{math_commands.tex}

\usepackage{hyperref}
\usepackage{url}
\usepackage{listings}
\usepackage{graphicx}
\usepackage{booktabs}
\usepackage{algpseudocode}
\usepackage{algorithm}
\usepackage{xspace}
\usepackage{multirow}
\usepackage{capt-of}
\usepackage[table]{xcolor}

\input{iclr2026/defs}

\title{PRISMM-Bench: A Benchmark of Peer-Review Grounded Multimodal Inconsistencies}

\author{
\makebox[\textwidth][c]{%
\textbf{Lukas Selch}$^{1}$ \quad
\textbf{Yufang Hou}$^{2}$ \quad
\textbf{M. Jehanzeb Mirza}$^{3}$ \quad
\textbf{Sivan Doveh}$^{4}$%
}\\
\makebox[\textwidth][c]{%
\textbf{James Glass}$^{3}$ \quad
\textbf{Rogerio Feris}$^{5}$ \quad
\textbf{Wei Lin}$^{1}$
}\\[0.5em]
$^{1}$Johannes Kepler University Linz \quad
$^{2}$Interdisciplinary Transformation University Austria \quad \\
$^{3}$MIT CSAIL \quad
$^{4}$Stanford University \quad 
$^{5}$ MIT-IBM Watson AI Lab\\
\website{https://da-luggas.github.io/prismm-bench} \url{https://da-luggas.github.io/prismm-bench/}\\
}

\iclrfinalcopy % Uncommented for the final version to show authors
\begin{document}

\maketitle

% {
% \blfootnote{
% $\dagger$ Correspondence: \tt\small{wlin2021at@gmail.com} }
% }

\input{iclr2026/sections/abstract}

\input{iclr2026/sections/introduction}

\input{iclr2026/sections/related_work}
\input{iclr2026/sections/method}

\input{iclr2026/sections/experiments}

\input{iclr2026/sections/conclusions}
\clearpage

\input{iclr2026/sections/ethics}
\bibliography{iclr2026_conference}
\bibliographystyle{iclr2026_conference}

\clearpage
\input{iclr2026/sections/appendix}

\end{document}

%% file: math_commands.tex
\usepackage{amsmath,amsfonts,bm}

\def\eqref#1{equation~\ref{#1}}
\def\1{\bm{1}}

\DeclareMathAlphabet{\mathsfit}{\encodingdefault}{\sfdefault}{m}{sl}
\SetMathAlphabet{\mathsfit}{bold}{\encodingdefault}{\sfdefault}{bx}{n}

%% file: iclr2026/defs.tex
\definecolor{darkergreen}{rgb}{0,0.4,0}

\newcommand{\myparagraph}[1]{\vspace{2pt}\noindent{\bf #1}}

\newcommand{\method}{PRISMM-Bench\xspace}

\makeatletter
\def\blfootnote{\gdef\@thefnmark{}\@footnotetext}
\makeatother

\newcommand{\website}[1]{%
  \href{#1}{\includegraphics[height=1.1em]{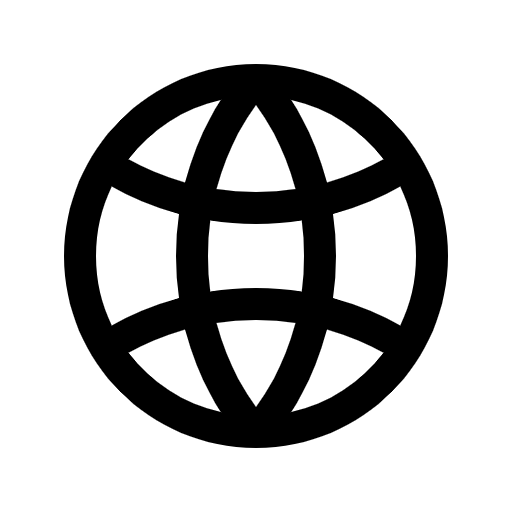}}%
}

\definecolor{jkuBlue}{RGB}{4,110,152}
\definecolor{jkuCyan}{RGB}{92,207,203}
\colorlet{jkuLightBlue}{jkuCyan}
\definecolor{jkuYellow}{RGB}{251,186,0}
\definecolor{jkuGrey}{RGB}{128,130,136}
\colorlet{jkuGray}{jkuGrey}
\definecolor{jkuDarkGrey}{RGB}{90,90,90}
\colorlet{jkuDarkGray}{jkuDarkGrey}
\definecolor{jkuLightGreen}{RGB}{187,208,50}
\definecolor{jkuGreen}{RGB}{76,172,78}
\definecolor{jkuPurple}{RGB}{156,71,123}
\definecolor{jkuRed}{RGB}{231,55,41}

%% file: iclr2026/sections/abstract.tex
\begin{abstract}

% \wei{alterative titles:\\
% PRISMM-Bench: A Multimodal Benchmark of Reviewer-Flagged Inconsistencies in Scientific Papers\\
% Lukas' version : PRISMM-Bench: A Multimodal Benchmark for Peer-Reviewed Visual Inconsistencies in Scientific Papers \\
% }

Large Multimodal Models (LMMs) are increasingly applied to scientific research, 
% supporting tasks such as figure interpretation, paper summarization, and error detection. 
yet it remains unclear whether they can reliably understand and reason over the multimodal complexity of papers. 
% In particular, detecting and resolving cross-modal inconsistencies such as misalignments between text, figures, tables, or equations remains a significant challenge because these issues are often subtle, require reasoning based on domain-specific knowledge, and ultimately undermine clarity, reproducibility, and trust. 
%A central challenge lies in inconsistencies, misalignments between text and figures, tables, or equations, which undermine clarity, reproducibility, and trust. 
A central challenge lies in detecting and resolving inconsistencies across text, figures, tables, and equations, issues that are often subtle, domain-specific, and ultimately undermine clarity, reproducibility, and trust.
Existing benchmarks overlook this issue, 
% either focusing on generic document QA or relying on synthetic inconsistencies that are tend to be obvious and unrepresentative of real-world complexity. 
either isolating single modalities or relying on synthetic errors that fail to capture real-world complexity.
We introduce PRISMM-Bench (\emph{\textbf{P}}eer-\emph{\textbf{R}}eview-sourced \emph{\textbf{I}}nconsistency \emph{\textbf{S}}et for \emph{\textbf{M}}ultimodal \emph{\textbf{M}}odels), the first benchmark grounded in real reviewer-flagged inconsistencies in scientific papers. 
Through a multi-stage pipeline of review mining, LLM-assisted filtering and human verification, we curate 384 inconsistencies from 353 papers.
% Our pipeline combines review mining, LLM-assisted filtering, and human verification, yielding 262 inconsistencies across 13 categories from 242 papers.  
%in 13 categories. 
Based on this set, we design three tasks, namely inconsistency identification, remedy and pair matching, which assess a model's capacity % detection, correction, and reasoning over 
to detect, correct, and reason over inconsistencies across different modalities. 
Furthermore, to address the notorious problem of \emph{choice-only shortcuts} in multiple-choice evaluation, where models exploit answer patterns without truly understanding the question, we further introduce structured JSON-based answer representations that  minimize linguistic biases by reducing reliance on superficial stylistic cues. 
We benchmark 21 leading LMMs, including large open-weight models (GLM-4.5V 106B, InternVL3 78B) and proprietary models (Gemini 2.5 Pro, GPT-5 with high reasoning). Results reveal strikingly low performance (27.8-53.9\%), underscoring the challenge of multimodal scientific reasoning and motivating progress towards trustworthy scientific assistants. %We provide the source code and dataset viewer in the appendix, and will release the full source code, dataset, and annotation tool publicly upon acceptance.

\input{iclr2026/fig_tex/teaser_figure}

\end{abstract}

%% file: iclr2026/fig_tex/teaser_figure.tex
\begin{figure}[h]
    \centering
    \includegraphics[width=\linewidth]{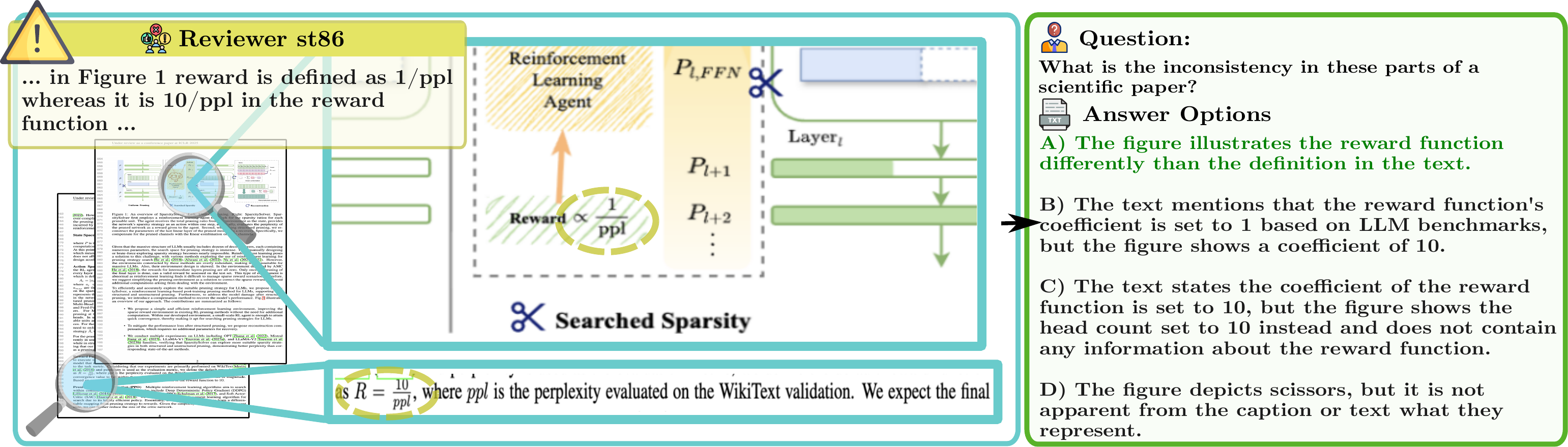}
    \caption{We collect reviewer-flagged inconsistencies in scientific papers and transform them into QA tasks that probe detection, correction, and reasoning over multimodal inconsistencies.}
    \label{fig:teaser}
\end{figure}

%% file: iclr2026/sections/introduction.tex
\section{Introduction}

% \wei{Teaser figure to show the concept of workflow}

Recent advances in Large Multimodal Models (LMMs) have sparked growing interest in their potential to serve as intelligent assistants for scientific research~\citep{eger2025transformingsciencelargelanguage}, supporting tasks such as figure and chart interpretation \citep{scifibench,tanasua2025rethinking,wu2024chartinsights}, paper summarization~\citep{tan2025enhancing,saxena2025postersum,yu2025deep}, and error detection \citep{yang2025can,miyai2024unsolvable,alsaif2024multimodal}. 
Yet, a central open question remains: \textit{can LMMs truly reason over the complex multimodal structure of scientific documents?}
% However, evaluating whether LMMs can truly reason over the multimodal structure of scientific documents remains an open challenge. 

A central challenge in this setting is detecting and resolving inconsistencies between text, figures, tables, or equations in scientific papers. 
These issues are often subtle, arising from copy-paste mistakes, outdated results, or inconsistent notation, and require domain knowledge to detect. Fig.~\ref{fig:teaser} illustrates such a case, where the reward function is defined differently in the figure and the in-line text.
Our analysis of ICLR 2025 submissions reveals that 17.0\% contained at least one such inconsistency flagged by reviewers. These discrepancies undermine clarity, reproducibility, and ultimately scientific trust.
% \footnote{We obtain this estimate by analyzing reviewer comments and identifying explicit mentions of inconsistencies across different modalities.}. 

% These issues are often subtle and require reasoning based on domain-specific knowledge, as illustrated in Figure \ref{}, in which \yufang{shortly describing the inconsistency problem that requires reasoning}. %Such inconsistencies typically arise when textual descriptions fail to align with supporting elements such as figures, tables, or equations, often due to copy-paste mistakes, outdated results, or versioning errors. 
% Such inconsistencies can arise not only between text and other supporting modalities, such as figures, tables, or equations, but also across these modalities themselves. They often stem from copy–paste mistakes, outdated results, versioning errors, or inconsistent notation. 
% %mismatched captions. %A critical challenge in this setting is the prevalence of inconsistencies within scientific papers.  
% Notably, among all ICLR 2025 submissions, 17.0\% \yufang{fill the number} contained one or more such inconsistencies flagged by reviewers\footnote{We obtain this estimate by analyzing reviewer comments and identifying explicit mentions of inconsistencies across different modalities.}.  
% Such discrepancies undermine clarity, reproduciblity, and ultimately scientific trust. 
% %\wei{refer to one inconsistency example in the teaser figure. }

Existing benchmarks, however, fall short of exposing this. 
%Some focus on generic document question-answering tasks~\citep{docvqa,infographicvqa,tatdqa} or on standalone scientific visual elements such as diagrams~\citep{dvqa}, charts~\citep{chartqa}, or tables~\citep{hitab}, which overlook the unique complexities of scholarly papers. 
% Others focus only on synthetic inconsistencies generated by LLMs~\citep{mmir} that are tend to be obvious and unrepresentative of real-world complexity. Neither approach reflects the subtlety or variety of errors found in real-world scientific works. 
Document QA datasets~\citep{docvqa,infographicvqa,tatdqa} or standalone scientific visual element tasks focusing on diagrams, charts, or tables~\citep{dvqa,chartqa,hitab} miss the multimodal dependencies of scholarly works.
Synthetic datasets~\citep{mmir} generate artificial errors, but these are often obvious and unrepresentative of real-world complexity. 
% Constructing a realistic benchmark is difficult: authentic inconsistencies are rare, scattered, and costly to verify.
% As a result, there are no systematic evaluation resources grounded in real and identified inconsistencies. 
% issues that reflect the true complexities of scientific communication. 
However, constructing such a benchmark of authentic inconsistencies is challenging as these cases are rare, scattered, and labor-intensive to verify, often requiring domain expertise to identify and validate. To address this, we leverage a valuable but underutilized resource - \textit{open peer reviews}, %where reviewers /%frequently 
focusing on instances where reviewers
flag mismatches 
%between text and accompanying elements such as figures, tables and equations, 
across different modalities, such as between text, figures, tables, and equations, 
thereby providing a natural source of real-world, human-identified inconsistencies. 
%\wei{mention that we use recent reviews from ICLR 2025 to avoid potential data contamination. the pipeline has the potential to support construction of a live benchmark. }

In this paper, we introduce \method, a \textit{\textbf{P}}eer-\textit{\textbf{R}}eview-sourced \textit{\textbf{I}}nconsistency \textit{\textbf{S}}et for \textit{\textbf{M}}ultimodal \textit{\textbf{M}}odels. Unlike synthetic datasets, \method captures inconsistencies explicitly flagged by human reviewers in scientific papers on OpenReview. Through a multi-stage pipeline combining large-scale review mining, LLM-assisted filtering, and rigorous human verification, we curate a dataset of 384 inconsistencies across 353 papers submitted to ICLR 2024 \& 2025, spanning 15 categories of visual-textual and inter-visual mismatches.
%By reflecting the real challenges of reading and verifying scientific papers, \method provides a principled resource for evaluting and improving LMMs in scholarly publishing. 
\method provides a principled resource for evaluating and improving LMMs, grounded in the real challenges of %reading and verifying 
understanding and verifying 
scientific papers. Grounded on these recent reviews, we minimize data contamination risk and demonstrate a pipeline with the potential to construct a continuously updated live benchmark. 
Building on this inconsistency set, we design a benchmark suite of three multiple-choice question (MCQ) tasks: 1) Inconsistency Identification - detect what the inconsistency is; 2) Inconsistency Remedy - determine how to fix the inconsistency; and 3) Inconsistency Pair Match - identify which two elements conflict. Together, these tasks form a tiered framework that evaluates not only a model’s ability to spot inconsistencies but also its capacity to propose remedy, and reason over relationships between different modality components. 

% \wei{mention that we use MCQ for eval before explaining this challenge of linguistic biases.}
In MCQ evaluation, a key challenge is models' tendency to rely on linguistic biases in answer choices.
% We use multiple-choice evaluation. However
% A further challenge is that multiple-choice evaluation is notoriously vulnerable to linguistic biases. 
Prior work has shown that LLMs often exploit choice-only shortcuts, achieving non-trivial accuracy without reading the question~\citep{answer_match, Gaming_truthfulqa, balepur2024artifacts,chizhov2025hellaswag}, 
and similar effects appear in multimodal MCQs~\citep{answer_match}. 
% Similar effects appear in multimodal MCQs where models can guess correctly without viewing the multimodal content~\citep{answer_match}. 
To address this, we propose a novel structured JSON-based answer representation that de-emphasizes stylistic cues and minimizes spurious correlations. 
Inspired by ~\cite{das-etal-2014-frame}~and ~\cite{banarescu-etal-2013-abstract}, our design converts free-form natural language into uniform structured representations that reduce model sensitivity to surface-level patterns. Our user study confirms that this approach suppresses linguistic shortcuts and better aligns models with human reasoning.

% This design choice was inspired by \emph{FrameNet-based semantic parsing} \citep{das-etal-2014-frame} and can be seen as  a form of \emph{abstract meaning representation} \citep{banarescu-etal-2013-abstract}, tailored to capture inconsistencies in scientific papers by  %\textit{authorship obfuscation}: 
% converting free-form natural language into uniform, structured representations that reduce model sensitivity to surface-level patterns.  Our user study confirms that our JSON formatting suppresses linguistic shortcuts and aligns model behavior more closely with human reasoning.
% \wei{mention the user study results to motivate this design choice. }

% Moreover, to mitigate the well-known issues of linguistic biases in multiple-choice evaluation, we further introduce structured JSON-based answer representations that significantly reduce reliance on superficial textual cues. 
% \wei{motivate the alleviation of linguistic biases via json format. mention this as a contribution. }

We benchmark 21 state-of-the-art LMMs, spanning large open-weight models such as GLM-4.5V 106B~\citep{glm4.5v} and InternVL3 78B~\citep{internvl3}, as well as leading proprietary models including Gemini 2.5 Pro~\citep{gemini2.5} and GPT-5~\citep{gpt5}. Results show that while large open-weight models achieve around 40\% accuracy, even the strongest proprietary models reach just 53.9\%, 
%underscoring the task difficulty and gap to human-level performance.
underscoring difficulty of the benchmark and limitations of current LMMs.
% show that our questions are quite challenging, large open-weight models at the range from 72B to 106B achieve only the performance of around 40\%, while the most competent proprietary multimodal models like Gemini 2.5 Pro and GPT-5 with high reasoning setting achieve only 54.2\%. 

Our contributions are fourfold: (1) We propose a reviewer-sourced dataset of real multimodal inconsistencies in scientific papers, spanning diverse categories and grounded in peer review.  (2) We construct a benchmark suite of three tasks probing detection, correction, and relational reasoning over these inconsistencies. (3) We are the first to propose JSON-based debiasing method for MCQ, converting free-form responses into uniform structured representations.
% structured debiasing for designing MCQ evaluation benchmarks via JSON-format answers, which convert free-form responses into uniform structured representations to mitigate language bias. 
(4) We evaluate 21 state-of-the-art LMMs, exposing their current limitations in detecting, understanding and correcting inconsistencies in scientific papers.
%The dataset and code for creating the benchmark and evaluating LMMs will be made publicly available upon acceptance.

% (3) we benchmark the performance of 16 state-of-the-art LMMs, including the most competent proprietary models like Gemini 2.5 Pro and GPT-5 with high reasoning setting, and Large Open-Weight models like GLM 4.5V 106B and InternVL3 78B which have high rankings on the multimodal leaderboard, exposing their weaknesses in terms of detecting, understanding and fixing inconstencies in scientific papers. 

%% file: iclr2026/sections/related_work.tex
\section{Related Work}

\myparagraph{Large Multimodal Models (LMMs).} 
Large Multimodal Models (LMMs) pair a vision encoder with a large language model, enabling open-ended reasoning across tasks such as image captioning, 
VQA, document understanding, and chart interpretation. Early approaches like BLIP-2~\citep{blip2} and InstructBLIP~\citep{instructblip} introduced instruction tuning on pre-trained vision-language models, while the LLaVA series~\citep{llava,llava+,llava_one_vision} advanced perception and reasoning via large-scale visual instruction tuning. 
Several recent studies~\citep{llava_icl,gavrikov2024vision,lin2024comparison,huang2024conme,glov,instructify,mei2025perla} have advanced these models by introducing improved training and adaptation strategies.
Recent models extend these capabilities: Qwen-2.5 VL~\citep{qwen2.5vl} offers precise object localization, dynamic resolution, and agentic tool execution; InternVL3~\citep{internvl3} improves perception and reasoning through domain-specific pretraining on 3D scenes, GUIs, and video; Gemma 3~\citep{gemma3}, Ovis 2~\citep{ovis2.5}, and GLM 4.5V~\citep{glm4.5v} demonstrate strong performance across diverse multimodal benchmarks. High-resolution variants such as InternLM XComposer 2.5~\citep{internlm_xc2.5} and VILA HD 4K~\citep{vila}, enable detailed perception and document processing. Proprietary models like GPT-5~\citep{gpt5} and Gemini 2.5 Pro~\citep{gemini2.5} set the state-of-the-art on complex multimodal tasks through large-scale training and enhanced reasoning.

% Overall, these models represent the current state of multimodal AI, forming the backbone for research in document reasoning, scientific understanding, and multimodal QA. Our work leverages this broad spectrum of LMMs to evaluate their capacity for detecting, understanding, and correcting inconsistencies in scientific papers.
These LMMs form the foundation for evaluating multimodal reasoning over scientific documents. In \method, we benchmark 21 top-performing models to detect, understand, and correct real-world inconsistencies in peer-reviewed papers, exposing both their strengths and limitations.

\myparagraph{Multimodal Benchmark on Scientific Paper Understanding.}
Prior benchmarks often focus on isolated scientific elements such as diagrams~\citep{dvqa,leafqa,figureqa}, charts~\citep{chartqa,plotqa}, or tables~\citep{hitab,fetaqa}. 
% Several benchmarks target QA on standalone scientific visual elements like diagrams~\citep{dvqa,leafqa,figureqa}, charts~\citep{chartqa,plotqa} and tables~\citep{hitab,fetaqa}. 
% DVQA~\citep{dvqa}, LEAF-QA~\citep{leafqa}, and FigureQA~\citep{figureqa} focus on diagrams, ChartQA~\citep{chartqa} and PlotQA~\citep{plotqa} on charts, and WTQ~\citep{wtq}, HiTab~\citep{hitab}, and FetaQA~\citep{fetaqa} on tables. 
Recent datasets like MathVista~\citep{mathvista}, MathVerse~\citep{mathverse}, and ArXivQA~\citep{arxivqa} integrate multiple modalities, but still treat figures and equations in isolation rather than in full-paper context. 
% For whole paper understanding, datasets such as PubMedQA~\citep{pubmedqa}, BioASQ~\citep{bioasq}, and QASPER~\citep{qasper} contain thousands of human-written questions, yet annotators typically only read abstracts when constructing them, leading to predominantly yes/no or short-span extractive questions.
Whole-paper QA resources such as PubMedQA~\citep{pubmedqa}, BioASQ~\citep{bioasq} and QASPER~\citep{qasper} provide human-written questions, yet these are mostly abstract-based and limited to yes/no or short-span answers.

Closer to our setting, QASA~\citep{qasa} provides 1.8K expert-written questions on ML papers, but remains text-only and does not require reasoning over figures or tables. SPIQA~\citep{spiqa} introduces multimodal scientific QA, yet the questions are LLM-generated or human-annotated with an emphasis on information seeking, not grounded on expert reviews that often aim to critically evaluate scientific findings. 
SciDQA~\citep{scidqa} is sourced from reviewer–author QA pairs, but it remains a text-only LLM benchmark without involving visual elements. In contrast, \method is the first benchmark grounded in reviewer-flagged multimodal inconsistencies in scientific papers. Unlike prior work that isolates figures, tables, or text, our benchmark integrates visual and textual reasoning within the natural context of full research papers, while grounding tasks in authentic peer review feedback rather than synthetic or abstract-level annotations.

% SCIDQA -  question-answer pairs from OpenReview, highly related 

% SPIQA – Scientific Paper Image QA (Google)

% SPIQA has a nice summary of datasets for QA on scientific papers

% SciFIBench – Scientific Figure Interpretation

% MME-SCI – Multilingual, Multimodal Scientific Benchmark

% SciVer – Scientific Claim Verification

% Evaluation of multimodal models on scientific paper understanding: SciVer~\cite{sciver}, SciFiBench~\cite{scifibench}, SPIQA~\cite{spiqa}

\myparagraph{Understanding of Inconsistencies.}
Research on inconsistencies in language models spans prediction variance across paraphrased queries~\citep{temple2020systematicity,consistency_lm} to factual inconsistency in summarization and long-form QA. 
% To address the latter, researchers have developed  QA-based benchmarks such as WikiContradict ~\citep{hou2024wiki}, evaluation metrics such as QAFactEval~\citep{qafacteval},   
%factual consistency QA~\citep{factual_consistency_qa}, %and recent 
%long-document approaches
% as well as methods based on  question answering~\citep{factual_consistency_qa}, natural language inference
% ~\citep{fast_inconsistency_detection} and  probabilistic inference~\citep{marinescu2025factreasonerprobabilisticapproachlongform}.
To address the latter, prior work has introduced QA-based benchmarks (e.g., WikiContradict~\citep{hou2024wiki}), evaluation metrics (e.g., QAFactEval~\citep{qafacteval}), and detection methods based on QA~\citep{factual_consistency_qa}, natural language inference~\citep{fast_inconsistency_detection}, or probabilistic reasoning~\citep{marinescu2025factreasonerprobabilisticapproachlongform}.

Closest to our setting, MMIR~\citep{mmir} evaluates multimodal models on artificially injected inconsistencies in materials such as slides and posters.
In contrast, \method introduces real-world reviewer-flagged inconsistencies in scientific papers. Rather than synthetic perturbations, our benchmark captures authentic challenges faced during scholarly review, spanning textual, visual, and cross-modal errors, and extends evaluation beyond detection to proposing remedies.

% By contrast, \method introduces reviewer-flagged inconsistencies from real scientific papers, capturing authentic textual, visual, and cross-modal errors. Unlike synthetic setups, it reflects real challenges in peer review and extends evaluation beyond detection to proposing remedies.

% MMIR  https://arxiv.org/abs/2502.16033 
% Multimodal inconsistency benchmark, but all the inconsistencies are synthetic, generated through webpage editing  \\
% LLM benchmarks 

\myparagraph{Language Biases in Evaluation Benchmarks.}
Multiple-choice evaluation is prone to linguistic biases, where models exploit surface-level patterns in answer options rather than reasoning over content. Prior studies show LLMs can achieve high accuracy even without the question, such as in TruthfulQA~\citep{Gaming_truthfulqa}, HellaSwag~\citep{hellaswag}, and ARC~\citep{balepur2024artifacts}. For example, \citet{answer_match} report 83\% accuracy on TruthfulQA v2 using answer choices alone, with shortcut rates above 70\% on HellaSwag. The recent trend of generating distractors with LLMs (e.g., in MMLU-Pro; \cite{wang2024mmluprorobustchallengingmultitask})
can even exacerbate these artifacts. %making choice-only heuristics more effective.
% making it even easier for models to exploit them. 

To mitigate such biases, structured representations offer a promising direction. Analogous to 
%abstract meaning representation (AMR) in NLP 
%or
authorship obfuscation in stylometry \citep{chinchor-1998-overview}, structured formats remove stylistic and surface cues while retaining semantics. Drawing inspiration from FrameNet-based semantic parsing \citep{das-etal-2014-frame} and MUC slot filling \citep{10.1145/3606274.3606276}, \method introduces JSON-based answer representations that encode key elements for capturing inconsistencies in scientific papers. This design reduces artifacts in answer choices and compels models to engage with multimodal content rather than exploiting linguistic shortcuts.

%% file: iclr2026/sections/method.tex
\input{iclr2026/fig_tex/pipeline_overview}

\section{\method}\label{method}

\method is built through a six-stage pipeline (Fig.~\ref{fig:pipeline}): (1) review sourcing, (2) LLM-based review filtering, (3) manual annotation of reviewer-flagged inconsistencies (Sec.~\ref{sec:collection_reviewer_flagged_inconsistencies}), (4) LMM-based task generation,  (5) manual verification to finalize benchmark tasks (Sec.~\ref{sec:generation_benchmark_tasks}), and (6) LLM-based debiasing to reduce language biases (Sec.~\ref{sec:alleviation_language_bias}). The evaluation step is introduced in Sec.~\ref{sec:eval_steup}.
%An overview of this pipeline is shown in Fig.~\ref{fig:pipeline}.

\subsection{Collection of Reviewer-Flagged Inconsistencies}\label{sec:collection_reviewer_flagged_inconsistencies}
To build a benchmark of realistic and authentic inconsistencies, %we avoid synthetic generation. Instead,
we sourced cases flagged by reviewers on OpenReview~\citep{openreview}, where comments often highlight discrepancies between textual content and visual or mathematical components, including figures, tables, and equations.

\myparagraph{Review Sourcing Strategy.}  
% \wei{make it clear that data sourcing means we download reviews, not the papers. we downnload all the reviews of ICLR 2025. All the ICLR 2025 reviews result in thes 262 inconsistency cases. }
We collected reviews from ICLR 2024 \& 2025 submissions via the OpenReview API v2~\citep{openreview_api}. 
% To maximize the chance of capturing unresolved inconsistencies, we focused on submissions that were rejected or withdrawn and had no author rebuttal. In such cases, inconsistencies flagged by reviewers are more likely to remain in the publicly available PDF rather than being corrected during the review process.
%Our earlier exploration of review sourcing showed that most reviewer-flagged inconsistencies were resolved during rebuttal and no longer appeared in the final versions, motivating this refined sourcing strategy. 
%\wei{use one sentence to describe how the data sourcing is done technically}
% This initial data sourcing resulted in 12,366 reviews. For more details of our review sourcing process, please refer to appendix \ref{app:data-coll}.
To maximize the likelihood that flagged inconsistencies persisted in the final public PDFs, we restricted to rejected or withdrawn submissions without rebuttals.\footnote{Our earlier exploration of review sourcing revealed that most reviewer-flagged inconsistencies were resolved during rebuttal and did not persist in the final versions, motivating the current refined sourcing strategy.}
This yielded 18,009 reviews (details in App.~\ref{app:data-coll}).

\myparagraph{LLM Review Filtering.} As manual screening 
%at this volume of 
for all reviews was infeasible, we employed an LLM for review filtering. 
Specifically, we used  \textit{Mistral Nemo 2407} \citep{mistral_nemo} with low temperature settings to summarize reviews and identify potential inconsistency mentions, resulting in a curated set of 6,056 potential inconsistencies spanning 2,458 reviews (prompt details in App.~\ref{app:prompt-review-filtering}).

\myparagraph{Manual Verification.} We performed a manual annotation pass using a custom web-based annotation tool. The tool presented the annotator with one reviewer-flagged inconsistency at a time, alongside the corresponding paper in an embedded PDF viewer. Annotators (1) verified whether reviewer comment described a factual and identifiable inconsistency, and (2) annotated the relevant textual and/or visual parts of the paper. For visual elements, the annotator selected and cropped regions from the PDF. For textual elements, they specified the page, line, and content. In addition, each inconsistency was assigned a category and a brief description in the annotator’s own words. The tool logged annotations together with the original reviewer’s comment and automatically collected metadata such as the crop bounding boxes in a structured format. Full details of the annotator background, annotation tool, captured metadata, annotation criteria and schema are provided in App.~\ref{app:ann-app}.

This process produced 384 validated inconsistencies across 353 ICLR submissions. We identified 15 categories of inconsistencies based on the elements involved (distribution shown in Fig.~\ref{fig:type-dist}). The most common cases were intra-figure inconsistencies (23.7\%) and figure-text mismatches (21.9\%). %\yufang{add numbers}

\subsection{Generation of Benchmark Tasks} \label{sec:generation_benchmark_tasks}

From the verified inconsistencies, we constructed three multiple-choice tasks with four options (one correct, three distractors), following the evaluation choice of most recent frontier model releases~\citep{qwen3,deepseekv3,gemini2.5,gemma2} and benchmarking efforts~\citep{mmlupro,zhang2025automated,livexiv}. We design the following three multiple-choice tasks.

% Building on the annotated inconsistencies, we designed three multiple-choice tasks to form the benchmark. 
% Following the evaluation choice of most recent frontier model releases~\citep{qwen3,deepseekv3,gemini2.5,gemma2} and benchmarking efforts~\citep{mmlupro,zhang2025automated}, 
% we opted for multiple-choice questions (MCQ) with four answer options (one correct, three distractors) to enable straightforward evaluation.
% and comparability with existing VQA benchmarks (e.g. see \cite{vqa}).
% We design the following three multiple-choice tasks. 

% \myparagraph{Inconsistency Identification.} The first task aims at evaluating the models' ability to identify the inconsistencies given the relevant context from the paper by asking the question \textit{``What is the inconsistency in these parts of a scientific paper?"}. Here we choose the question of generic style as we realized in a preliminary study that sample-specific questions like \textit{``What inconsistency is observed between Figure 2 and the accompanying text regarding the generated road network?"} make the task too easy. 
\myparagraph{Inconsistency Identification (\textit{Ident}).} The first task evaluates a model's ability to recognize inconsistencies within the given paper context, framed by the question: \textit{``What is the inconsistency in these parts of a scientific paper?"} We adopt this generic question style because our preliminary study showed that sample-specific questions (e.g.  \textit{``What inconsistency is observed between Figure 2 and the accompanying text regarding the generated road network?"}) reveal the inconsistency content and oversimplify the task (see App.~\ref{app:debiasing_inconsistency_identification} for details). 

% To construct candidate answers, we instructed \textit{Google Gemini 2.5 Flash} to generate options based on the annotated inconsistency descriptions and the corresponding visual/textual elements. 
Candidate answers were generated using \textit{Gemini 2.5 Flash} based on inconsistency descriptions and corresponding multimodal context. 
% We instructed \textit{Google Gemini 2.5 Flash (Gemini Flash)} to generate candidate answers from the annotations, given the visual/textual parts and the annotator's description of the inconsistency. 
The answers were manually refined to ensure (1) the correct choice captured the inconsistency precisely and (2) the distractors are contextually relevant and plausible, but incorrect. We show an example of the \textit{Ident} task in Fig.~\ref{fig:pipeline}.

\myparagraph{Inconsistency Remedy (\textit{Remedy}).} This task extends beyond simple detection by requiring models to how to fix the inconsistency by asking the question \textit{``What action needs to be taken to resolve the inconsistency in these parts of a scientific paper?"} \textit{Gemini 2.5 Flash} was employed to reformulate the inconsistency statements from \textit{Ident} into specific, actionable remedy formulations. This task evaluates whether models can propose plausible solutions, rather than merely spotting inconsistencies. 

% \myparagraph{Inconsistency Pair Match.} Here we focus on a subset of inconsistencies composed of two distinct visual elements in the paper (135 samples). Models are given the one element as the context and are instructed to pick the correct element out of the four answer options with which a visual inconsistency is formed. By isolating inconsistencies to visual-only pairs, we can focus on assessing a model’s ability to handle data representation errors without relying on textual cues, simulating the common peer review challenge of cross-referencing figures and tables for consistency.
\myparagraph{Inconsistency Pair Match (\textit{Match}).} This task is built on a subset of inconsistencies that involve two distinct visual elements within a paper (192 samples). 
Given one element as context, the model must select its inconsistency counterpart from four options.
% the element that forms an inconsistent part. 
By restricting the task to visual-visual mismatches, we specifically assess a model's ability to detect representation errors without relying on textual cues, simulating the common peer-review challenge of cross-checking figures and tables for consistency. 
% An overview of the three tasks is provided in \autoref{fig:example-datapoint} and 

More details about the task generation process and a validation of the multiple-choice format against open-ended evaluation are provided in App.~\ref{app:task-generation} and App.~\ref{app:mcq-vs-open-ended}, respectively. 
We provide qualitative examples of the three tasks in App.~\ref{app:dataset-viewer} and a dataset viewing tool in the supplementary materials. 
% In the supplementary materials to this paper we provide a dataset viewer application (see Appendix~\ref{app:dataset-viewer}) to explore more qualitative examples.
% \wei{mention that we have a dataset viewer in the supplementary materials (or refer to more qualitative examples if we have them in appendix) } 

% \input{iclr2026/fig_tex/inconsistency_task_examples}

\subsection{Alleviation of Linguistic Biases} \label{sec:alleviation_language_bias}

During pilot experiments, we observed that models achieved well above random accuracy even when the visual context was withheld. For example, \textit{Gemini 2.5 Flash} reached 57.6\% accuracy on the \textit{Ident} task without context (vs. 25\% random chance).
% , compared to a random baseline of 25\%. 
This indicated that models exploit linguistic priors and surface patterns in the answer options rather than reasoning over the actual content. Further analysis showed that factors such as answer length, relative position, and phrasing contributed to this bias, echoing known challenges in multiple-choice design \citep{mcq_design_review}. %\wei{mention the user study results to motivate the design choice}

To combat this bias, we first tried refining the distractors with text manipulation, which proved insufficient. Therefore, we introduced structured representations that minimize natural language cues. We designed the \textit{Evidence–Claim JSON} format for the \textit{Ident} task and the \textit{Target–Action JSON} format for the \textit{Remedy} task. Converting answers into these structured formats using an LLM reduced \textit{Gemini 2.5 Flash's} no-context accuracy on the \textit{Ident} task to 34.0\%. We manually verified a 20\% subset of the inconsistencies to ensure the semantic fidelity of the JSON-formatted answers. 
% remained consistent with their original counterparts. 
An example of the Evidence-Claim JSON format for the \textit{Ident} task is provided in Fig.~\ref{fig:pipeline}. 
Full details on our debiasing procedure and the structured formats are provided in App.~\ref{app:debiasing_inconsistency_identification}. 

This design choice is further supported by our user study (Sec.~\ref{sub:user-study}), which reveals that humans rely minimally on linguistic priors. In contrast, models evaluated on natural language options maintain high accuracy without context, exposing a fundamental mismatch in evaluation fidelity. By adopting structured JSON representations, we align model evaluation conditions more closely with human cognitive constraints, suppressing surface-level shortcuts and enabling a fairer assessment of true multimodal reasoning.

\subsection{Contextual Granularity in Evaluation}\label{sec:eval_steup}
We evaluate model performance under three levels of contextual granularity, reflecting different real-world reading conditions and reasoning demands.
% \paragraph{Metric} We adopt \textbf{accuracy} as the primary evaluation metric across all tasks and modalities. Given the multiple-choice format with four options (A-D), random guessing yields a baseline accuracy os 25\%. We define an answer as correct if the letter predicted by the model matches the letter of the correct answer.

% \paragraph{Contextual Granularity}

\myparagraph{Focused Context (\textit{Focused}).} The model is presented only with the minimal necessary components — an extracted visual element (e.g., cropped figure or table) and/or the precise text passage (e.g., sentence or paragraph) involved in the inconsistency, as annotated. This setting isolates the key content, testing the model’s ability to detect inconsistencies with minimal noise.
% maximal signal-to-noise ratio.

\myparagraph{Page Context (\textit{Page}).} The model receives a 144 DPI rasterized image of the entire page(s) where the inconsistency occurs. Visual elements are not pre-cropped, requiring the model to locate and interpret relevant content within the full page layout. 
% ; the model must locate and interpret relevant content within the page layout. 
This simulates realistic reading conditions where inconsistencies must be identified without prior localization.

\myparagraph{Document Context (\textit{Document}).} The model receives the entire scientific paper as a sequence of page images. To accommodate architectural constraints, we follow MMLongBench-Doc \citep{mmlongbenchdoc} and segment the document into collages: 
a total of 5 images are fed to the model, each containing $n_{pages}/5$ pages arranged in a 3-column grid. 
% We feed a total of 5 images to the model at once, each image containing $n_{pages}/5$ pages per image concatenated in a grid of three columns. 
This setting evaluates the model’s capacity for long-range, cross-page reasoning and document-level grounding. 
% One notable deviation from this setting are the models \textit{LLaVA Onevision (7B, 72B)}, which, due to their AnyRes high-resolution processing strategy, faced token limitations and required a reduced input of 3 images containing $n_{pages}/3$ pages per image to avoid exceeding the context window.
For models with high-resolution processing constraints, such as \textit{LLaVA Onevision (7B, 72B)}, we reduce input to 3 images with $n_{pages}/3$ pages each to avoid exceeding the context window. 

% \begin{enumerate}
% \item \textbf{Focused Context (Focused)}: The model is presented only with the minimal necessary components—an extracted visual element (e.g., cropped figure or table as PNG) and/or the precise text passage (e.g., sentence or paragraph) involved in the inconsistency, as annotated. This setting tests the model’s ability to detect inconsistencies with maximal signal-to-noise ratio.

% \item \textbf{Page Context (Page)}: The model receives a 144 DPI rasterized image of the entire page(s) where the inconsistency occurs. Visual elements are not pre-cropped; the model must locate and interpret relevant content within the page layout. This simulates real-world reading conditions where inconsistencies must be identified without prior localization.

% \item \textbf{Document Context (Document)}: The model is given the complete scientific paper as a sequence of page images. To manage architectural constraints, we align with MMLongBench-Doc \citep{mmlongbenchdoc} by segmenting the document into collages: We feed a total of 5 images to the model at once, each image containing $n_{pages}/5$ pages per image concatenated in a grid of three columns. This setting evaluates the model’s capacity for long-range, cross-page reasoning and document-level grounding. One notable deviation from this setting are the models \textit{LLaVA Onevision (7B, 72B)}, which, due to their AnyRes high-resolution processing strategy, faced token limitations and required a reduced input of 3 images containing $n_{pages}/3$ pages per image to avoid exceeding the context window.
% \end{enumerate}

%% file: iclr2026/fig_tex/pipeline_overview.tex
\begin{figure}[h]
\begin{center}
\includegraphics[width=\linewidth]{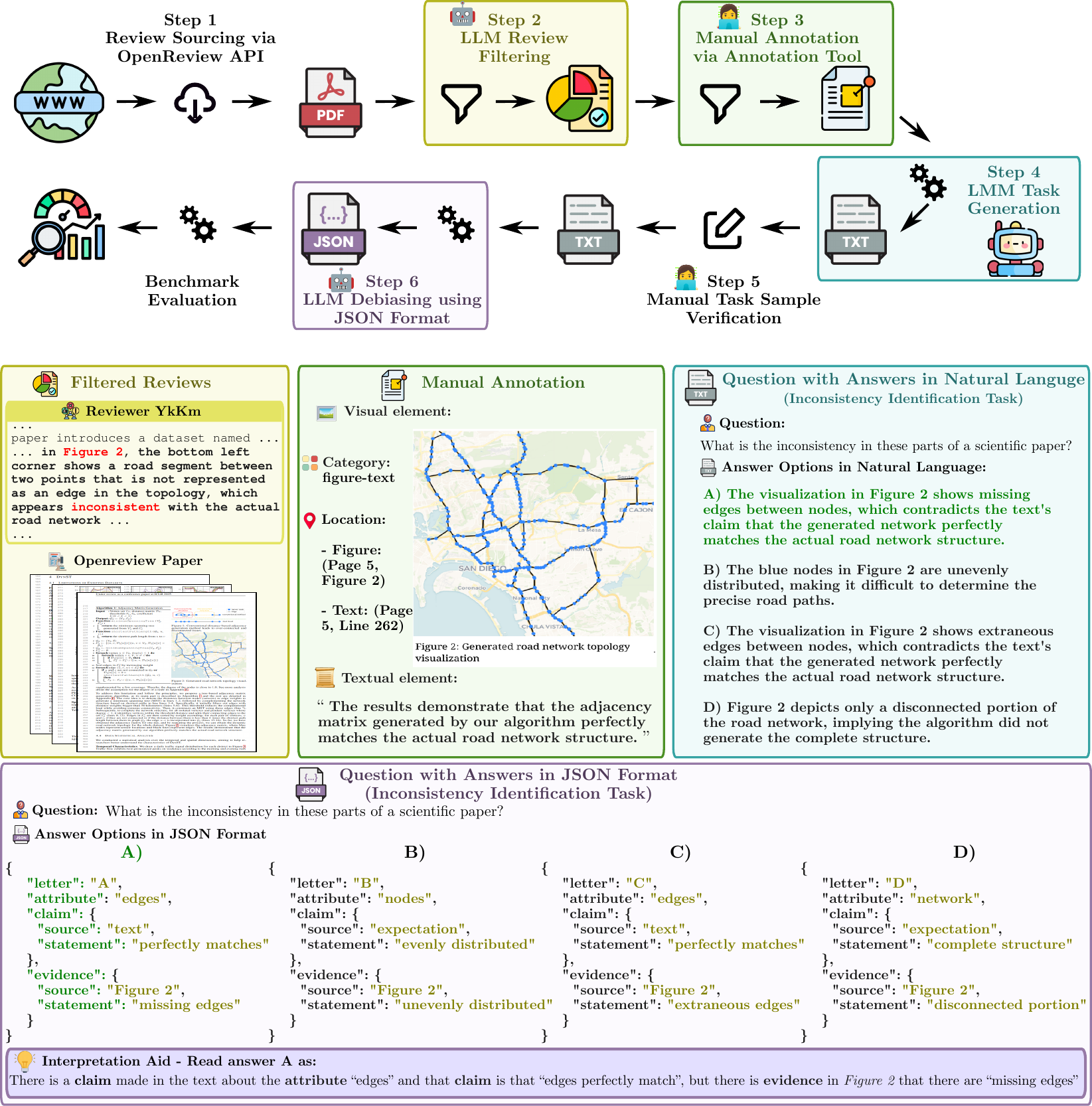}
\end{center}
\vspace{-0.3cm}
\caption{\textbf{Pipeline of \method.} 
% Our pipeline (top pipeline flow) involves collection of reviewer-flagged inconsistencies (step 1 review sourcing, step 2 LLM-based filtering and step 3 manual annotation, see Sec.~\ref{sec:collection_reviewer_flagged_inconsistencies}), generation of benchmark tasks (step 4 LLM question generation and step 5 manual question verification, see Sec.~\ref{sec:generation_benchmark_tasks} ) and alleviation of language biases (step 6 LLM debiasing, see Sec.~\ref{sec:alleviation_language_bias}). 
% At the bottom of the figure, We show the filtered reviews after step 2, example annotation result after step 3, example question with answers in natural language after step 4, and example question with answers in JSON format after step 6. 
% Our approach involved the close collaboration of LLM-automated editing and human annotation.   \wei{xxx } 
The top row illustrates the six main steps: (1) review sourcing, (2) LLM-based review filtering, and (3) manual annotation of metadata for reviewer-flagged inconsistencies (Sec.~\ref{sec:collection_reviewer_flagged_inconsistencies}), (4) LMM-based task  generation, (5) manual verification to construct benchmark tasks (Sec.~\ref{sec:generation_benchmark_tasks}), and (6) LLM-based debiasing to mitigate language biases (Sec.~\ref{sec:alleviation_language_bias}). The bottom row shows representative outputs at each stage: filtered reviews after step 2, inconsistency annotation after step 3, an example multiple-choice question in natural language after step 4, and its debiased JSON-format counterpart after step 6.
}
\vspace{-0.3cm}
\label{fig:pipeline}
\end{figure}

%% file: iclr2026/sections/experiments.tex
\section{Experiments}

\subsection{Experimental Setup}

\myparagraph{Model Selection.} 
% To ensure a comprehensive and fair evaluation, we curated a diverse set of 21 LMMs, including 16 open-weight models of varying scales, two specialized high-resolution models, and three proprietary models. 
We evaluate 21 LMMs spanning a diverse range of architectures: 16 open-weight models of varying scales, two specialized high-resolution models, and three proprietary models.
Selection was guided by performance on the Open VLM Leaderboard~\citep{open_vlm_leaderboard} and the availability of the latest model versions.
% or are more recent versions of those. 
% Additionally, we included two specialized high-resolution models explicitly designed for fine-grained visual understanding, as well as two leading proprietary models to establish performance upper bounds.

\myparagraph{Inference Details.} 
% To minimize noise from model outputs and ensure consistent scoring, we enforce a strict answer format via prompting the models in both a system-level and user-level prompt to explicitly output only the letter corresponding to the predicted correct answer. We post-process the answers by removing any leading and trailing whitespace, as well as special characters. For reasoning models, we first separate the reasoning within \texttt{<think></think>} tags from the actual answer before proceeding with the extraction and matching process.
To ensure consistent scoring, we enforced a strict answer format. 
% Models were instructed, via both system- and user-level prompts, to output only the letter of the chosen option, followed by our postprocessing. 
Prompts at both the system and user level instructed models to output only the letter corresponding to their chosen option.
% For reasoning-enabled models, we extracted the answer separately from the reasoning, which was wrapped in \texttt{<think></think>} tags, before matching against the ground truth. 
For reasoning-enabled models, answers were extracted separately from reasoning traces, enclosed in \texttt{<think></think>} tags, before postprocessing against the ground truth.

All open-weight models were grouped into three parameter count categories and evaluated with \texttt{vLLM} v0.10.1~\citep{vllm}. Experiments were conducted on 4$\times$NVIDIA A100 64GB GPUs with greedy decoding, except for InternVL3.5 (8B, 38B) which required a temperature of 0.6 for stable reasoning. Proprietary models were accessed via their official APIs with greedy decoding except \textit{GPT-5 (minimal, high)} which has a fixed temperature of 1.0. 
% where the recommended temperature of 0.6 was needed for successful reasoning.

% For proprietary models, we used their official API with greedy decoding with the exception of \textit{GPT-5 (minimal, high)}, which has a fixed temperature setting of 1.0.

Each model was evaluated on all three tasks (\textit{Ident}, \textit{Remedy}, \textit{Match}), and across three contextual granularity levels (Sec.~\ref{sec:eval_steup}). 
% For \textit{Match}, the \textit{Page Context} and \textit{Document Context} settings were excluded. 
For \textit{Match}, only \textit{Focused Context} was used.
This design yields seven evaluation configurations per model, enabling fine-grained analysis of how model architecture, scale and input context affect inconsistency detection performance. 
% This resulted in a total of 7 evaluation settings per model, enabling fine-grained insights into how model architecture, size, and input context jointly influence inconsistency detection performance.

% \myparagraph{Answer Extraction.} 

\subsection{Main Results}

\input{iclr2026/tables/main_results}

% Our evaluation across a diverse set of LMMs reveals key finding regarding their capabilities in understanding and correct inconsistencies in scientific documents. The aggregated results are presented in Table~\ref{table:bench_results}.
% Our evaluation across a diverse set of LMMs reveals key insights into their ability to detect and correct inconsistencies. Aggregate results are shown in Table~\ref{table:bench_results}.
Table~\ref{table:bench_results} summarizes the aggregate performance of all evaluated models. Our results reveal clear trends in how LMMs handle inconsistency detection and correction.

% \myparagraph{Performance Landscape.} Proprietary models substantially outperform their open-weight counterparts. On our benchmark, the top proprietary models (Gemini 2.5 Pro, GPT-5 high) have a state-of-the-art accuracy of 54.2\%. In contrast, the top open-weight model (GLM 4.5V 106B) scores 41.9\%. Although it matches the performance of the minimal-reasoning GPT-5 in most tasks, it still lags behind the high-reasoning version by 12.3 points.

% The overall scores highlight the challenge the LMMs face in the role of a trustworthy scientific assistant. As even the best models' performance are strikingly low, more work has to be done in order to employ multimodal models as automated peer-review agents.

\myparagraph{Performance Landscape.} Proprietary models substantially outperform their open-weight counterparts. GPT-5 (high reasoning) reached the highest average performance of 53.9\%. By contrast, the best open-weight model GLM 4.5V 106B achieves 42.5\%, matching GPT-5 (minimal reasoning) but trailing its high-reasoning variant by 11.4 points. These results underscore the difficulty of the benchmark: even the best-performing models remain far from the reliability required of automated scientific assistants. 

% \myparagraph{Impact of Context and Task Formulation.} 
% We observe a clear trend related to the context scope. For most of the models, performance consistently degrades as the context expands. Models of all sizes achieve their highest scores in the \textit{Focused} setting, with the accuracy often hovering close to random chance for the \textit{Full Pages} and \textit{Full Document} context. This suggests models struggle with distraction and long-range information grounding in dense, multi-page contexts.

\myparagraph{Impact of Context and Task Formulation.} 
% Context granularity has a strong effect on performance. Across almost all models, accuracy drops as the input scope expands: 
Performance drops consistently as context scope expands.
% results are highest in the \textit{Focused} setting, but often approach random chance under \textit{Page} or \textit{Document} settings. This highlights persistent challenges with distraction and long-range grounding in dense, multi-page inputs.  
Models achieve their best accuracy in the \textit{Focused} setting but often degrade toward random chance under \textit{Page} and \textit{Document} inputs, reflecting persistent challenges with distraction and long-range grounding in dense, multi-page inputs.
To rule out input quality effects, we performed an ablation study on rasterization resolution, confirming our 144 DPI baseline (cf. App.~\ref{app:dpi}).
% To ensure our evaluation was not confounded by input quality, we performed an ablation study on rasterization resolution, confirming our 144 DPI baseline (cf. App.~\ref{app:dpi}).
% The formulation of the task itself strongly influences performance. The \textit{Remedy} task consistently yields lower accuracies than the \textit{Ident} task. This supports the hypothesis that correcting an error demands a deeper level of reasoning beyond simple detection. Performance on the \textit{Pair Match} task varies widely across different models, pointing to significant architectural differences. For example, Gemma 3 12B excels at this task with an accuracy of 64.4\%, pushing into proprietary model performance. Conversely, much larger models like InternVL3 78B (45.9\%) that outperform smaller-sized models on nearly ever task show weaknesses at this task, suggesting certain models possess stronger relational reasoning within a document structure than others.

Task formulation also plays a critical role. 
% The \textit{Remedy} task consistently yields lower scores than \textit{Ident}, confirming that processing corrections requires deeper reasoning than simple detection. 
\textit{Remedy} scores are consistently lower than \textit{Ident}, showing that proposing corrections requires deeper reasoning than only detection.
% Performance on \textit{Match} varies widely across models, reflecting architectural differences in relational reasoning. Notably, Gemma 3 12B achieves 64.4\% on this task, rivaling proprietary models, while much larger models like InternVL3 78B (45.9\%) underperform, suggesting that scale alone does not guarantee robust structural reasoning. 
Performance on \textit{Match} varies widely across models: Gemma 3 12B achieves 63.5\%, rivaling proprietary models, whereas much larger models such as InternVL3 78B lag behind at 45.3\%. These results suggest that architectural design, not just scale, is critical for relational reasoning.

% \myparagraph{Analysis of Model Characteristics.} Most evident is the overperformance of reasoning-enabled models. This is most evident when comparing InternVL3.5 8B to it's non-reasoning predecessor InternVL3 8B. The reasoning version achieves an average accuracy 5 percentage points higher than any other model in it's parameter count class, matching the performance of models with nine times higher parameter count. This demonstrates the clear benefit of reasoning for these complex tasks. Models specifically trained for high-resolution images (VILA HD 4K 8B, InternLM XC 2.5 7B) do not perform better on the extended context settings. Our findings challenge the "bigger is better" paradigm of model scaling. Simply increasing the parameter count does not guarantee improved performance. Particularly, we can see diminishing returns scaling from medium-sized to large-sized models.

\myparagraph{Model Characteristics.} Reasoning-enabled models show benefit. For example, InternVL3.5 8B outperforms its non-reasoning predecessor InternVL3 8B by over 5 percentage points, achieving accuracy comparable to models with nearly nine times more parameters. 
% This advantage is directly attributable to reasoning capabilities — our ablation shows that turning off chain-of-thought degrades performance substantially, with some models losing over 19 percentage points (cf. App.~\ref{app:cot}).
Turning off chain-of-thought reduces accuracy by up to 14 points (cf. App.~\ref{app:cot}), directly confirming the contribution of reasoning. 
%This highlights the benefit of reasoning capabilities for complex inconsistency detection tasks. 
% In contrast, models specialized for high-resolution images (VILA HD 4K 8B, InternLM XC 2.5 7B) do not yield improvements under extended context settings. 
In contrast, high-resolution specialists (VILA HD 4K 8B, InternLM XC 2.5 7B) show little advantage in extended-context settings.
More broadly, our results challenge the ``bigger is better" paradigm: scaling up parameter counts alone does not guarantee higher performance, with diminishing returns observed from medium- to large-scale models. 

Overall, these findings highlight the current limitations of LMMs for scientific document analysis. 
% Improved reasoning architectures are needed to scale beyond detection towards correction of inconsistencies and mechanisms for robust grounding over long and distractive context need to be established.
Future progress will require advances in reasoning architectures to move beyond error detection toward correction, as well as more robust mechanisms for grounding over long, distractive contexts. 

%\subsection{User Study}
\subsection{Analysis of Chain-of-Thought Reasoning}
\label{sec:cot}

Reasoning variants perform better than non-reasoning counterparts and reach results comparable to much larger models. For instance, InternVL3.5 8B achieves an average of 37.7\%, rivaling large open-weight models and surpassing several 72B non-reasoning models. We therefore study how reasoning-enabled models leverage chain-of-thought (CoT) to improve performance on \method. To do so, we re-evaluated a selection of reasoning models on the \textit{Ident} task with \textit{Focused Context} with reasoning turned off and compared the performance.

\myparagraph{Ablation Results.} Disabling reasoning reveals the critical role of CoT in detecting subtle inconsistencies. For example, GLM 4.5V drops from 51.8\% to 43.2\% (-16.6\%), InternVL3.5 8B from 49.5\% to 40.6\% (-18.0\%), and InternVL3.5 38B suffers the largest decline, from 54.4\% to 40.4\% (-25.7\%).

\myparagraph{Why Reasoning Helps.} To understand these performance differences, we focused on inconsistencies where InternVL3.5 38B succeeded with reasoning but failed without.
We identified three consistent patterns: (1) Structured input handling: Reasoning-enabled models interpret the JSON-formatted options in natural language, clarifying subtle distinctions without exploring linguistic biases (cf. low without-context performance for reasoning models in Sec.~\ref{par:ling-bias}). (2) Cross-modal grounding: CoT traces show models explicitly reasoning over both text and visuals, breaking complex information into smaller units and reusing them later in the reasoning chain. (3) Concept linking: Reasoning enables models to connect fine-grained context with domain knowledge and abstract concepts, allowing stronger logical inference beyond surface pattern recognition. We provide a detailed case study illustrating these effects in App.~\ref{app:cot}.

\subsection{User Study and Linguistic Bias Analysis}
\label{sub:user-study}

To complement our benchmark, we conducted a user study to establish a human baseline and quantify \textit{visual reliance} — the extent to which answers depend on genuine multimodal reasoning rather than linguistic shortcuts. While our benchmark uses structured JSON outputs for models, our participants are evaluated on natural language (NL) questions as structured formats are less practical without prior training. To enable direct comparison, we re-evaluated representative LMMs on the same \textit{Ident} task subset using natural language answer options.
% as our participants using natural language answer options, enabling direct comparison.

\myparagraph{Setup.} Eight non-author participants with at least PhD-level computer science research experience each answered ten randomly sampled \textit{Ident} task questions: five in \textit{Focused} context and five in \textit{Document} context. We provide a detailed analysis of the representativeness of this subset in App.~\ref{app:user-study-represent}.
For each question, participants first answered without context (question + answer options only), then with context. \textit{Focused} context consisted of cropped images and/or text excerpts; \textit{Document} context contains links to original PDFs. The survey was implemented via a custom web interface (cf. App.~\ref{app:user-study}).

\input{iclr2026/tables/user_study_table}

\myparagraph{Analysis.} 
As shown in Table~\ref{table:user-study}, top models like Gemini 2.5 Pro exceed humans under \textit{Focused Context} (81.6\% vs. 77.5\%) and \textit{Whole Document Context} (83.8\% vs. 65.0\%).
% While LMMs appear competitive with human reasoning, a closer look at the "without context" condition reveals a key difference. LMMs maintain high accuracies (up to 71.2\%), whereas human performance approaches random chance (35.0\%). This gap suggests that LMMs are exploiting linguistic regularities, while human reasoning is less reliant on these biases.
However, a crucial difference emerges in the \textit{Without Context} condition: LMMs maintain high accuracy (up to 70.1\%), whereas human performance drops near chance (27.5\%). This indicates that LMMs rely heavily on linguistic regularities that humans cannot exploit. 
Switching to JSON formatting neutralizes this advantage. Without context, model performance collapses toward human levels (e.g., InternVL3.5 38B drops from 53.7\% to 25.3\%). With context and JSON-structured answer, LMMs no longer match human NL performance, confirming that linguistic shortcuts inflate perceived model capability. 
% This demonstrates that when linguistic shortcuts are suppressed, LMMs fall behind humans in true visual grounding.

To quantify how much models and human rely on visual evidence versus linguistic priors, we compute the \textbf{Visual Reliance Ratio} $R$, adapted from the normalized Perceptual Score \citep{gat2021perceptualscore}:
\begin{equation} \label{eq:vis-reliance}
R = \frac{Acc_{\text{with\_context}} - Acc_{\text{without\_context}}}{1 - Acc_{\text{without\_context}}}
\end{equation}
% This metric isolates the relative gain from visual context: 
Higher $R$ indicates stronger dependence on visual context. Human participants achieve $R=69.0\%$, while the top model (InternVL3.5 8B) achieves $R=53.5\%$, confirming that humans rely more on visual grounding than current LMMs. 
% while the highest value for our LMM selection (InternVL3.5 8B) saw a relative gain of 46.8\%. 
% Our human participants therefore relied more heavily on visual grounding compared to the models.

\myparagraph{Probing Linguistic Bias.}\label{par:ling-bias} To confirm this effect generalizes beyond the user study subset, we re-evaluated the same four representative LMMs on the full \textit{Ident} dataset under both Natural Language and JSON formats (Table~\ref{table:nocontext-ablation}). The same pattern holds: under natural language, models achieve inflated accuracies without context (e.g. 61.2\% for Gemini 2.5 Pro) but performance drops toward chance under JSON. 
% but scores decrease closer to chance when forced into structured JSON. 
Correspondingly, $R$ increases under JSON for all models, showing that structured outputs suppress linguistic shortcuts and force models to rely more on visual evidence.

\input{iclr2026/tables/no_context_ablation_table}

\myparagraph{Insights.} Two key conclusions emerge: (1) MCQs with long-form answers in natural language overstate LMM performance, as models can exploit linguistic regularities imperceptible or irrelevant to humans.  
% by allowing exploitation of linguistic regularities imperceptible or irrelevant to humans. 
(2) Structured JSON representations mitigate this bias, revealing that even the strongest LMMs still fall short of human-level visual grounding and rely on surface cues when available.

%% file: iclr2026/tables/main_results.tex
\begin{table*}[ht]
\centering
\caption{
% \textbf{Benchmark results.} 
% We report the accuracy as fraction of correct answer after post-processing. Models marked with $^{\textbf{R}}$ are reasoning models. 
Accuracy (\%) of 21 diverse LMMs across three tasks (\textit{Ident}, \textit{Remedy}, \textit{Match}) and three levels of contextual granularity (Sec.~\ref{sec:eval_steup}). For \textit{Match}, results are reported only under the \textit{Focused} setting. %, excluding \textit{Page} and \textit{Document} contexts. <-- commented this out to prevent 10th page
Best result per task bolded, second best underlined. $^{\textbf{R}}$ denotes reasoning models.
}
\label{table:bench_results}
\small
\resizebox{\linewidth}{!}{
\begin{tabular}{@{}lc|ccc|cc|cc|c@{}}
\toprule
 \multirow{3}*{\textbf{Model}}  & & \multicolumn{3}{c|}{\textbf{Focused}} & \multicolumn{2}{c|}{\textbf{Page}} & \multicolumn{2}{c|}{\textbf{Document}} & \\
\cmidrule(lr){3-5} \cmidrule(lr){6-7} \cmidrule(lr){8-9}
\textbf{} & \textbf{Params} & \textbf{Ident} & \textbf{Remedy} & \textbf{Match} & \textbf{Ident} & \textbf{Remedy} & \textbf{Ident} & \textbf{Remedy} & \textbf{Average} \\
~ & \textbf{[B]} & (384) & (384) & (192) & (384) & (384) & (384) & (384) & (960)\\
\midrule
% \textcolor{gray}{\textit{Random Chance}}  & - & \textcolor{gray}{25.0} & \textcolor{gray}{25.0} & \textcolor{gray}{25.0} & \textcolor{gray}{25.0} & \textcolor{gray}{25.0} & \textcolor{gray}{25.0} & \textcolor{gray}{25.0} & \textcolor{gray}{25.0}\\
% \midrule
\multicolumn{10}{c}{\textit{Small Open-Weight Models (\textless 9B)}} \\
\midrule
Gemma 3 4B & 4.0 & 27.9 & 29.9 & 39.6 & 25.0 & 24.7 & 26.6 & 27.1 & 27.8 \\
LLaVA OV 7B & 7.0 & 30.5 & 28.4 & 29.7 & 32.0 & 28.4 & 28.1 & 27.9 & 29.2 \\
Ovis2 8B & 8.0 & 35.4 & 29.4 & 22.4 & 34.4 & 27.3 & 31.8 & 28.1 & 30.4 \\
Qwen 2.5 VL 7B & 7.0 & 32.8 & 31.3 & 58.9 & 29.9 & 29.7 & 26.8 & 27.1 & 31.9 \\
InternVL3 8B & 8.0 & 36.5 & 29.4 & 56.3 & 28.6 & 27.6 & 30.7 & 31.8 & 32.7 \\
InternVL3.5 8B$^{\textbf{R}}$ & 8.0 & 49.5 & 35.9 & 45.8 & 38.3 & 30.5 & 36.7 & 31.0 & 37.7 \\
\midrule
\multicolumn{10}{c}{\textit{Medium Open-Weight Models (9B—38B)}} \\
\midrule
Gemma 3 27B & 27.0 & 36.2 & 32.8 & 59.9 & 30.7 & 28.6 & 31.0 & 27.3 & 33.3 \\
Gemma 3 12B & 12.0 & 33.9 & 30.5 & 63.5 & 30.7 & 25.8 & 30.7 & 30.5 & 32.9 \\
Qwen 2.5 VL 32B & 32.0 & 42.4 & 37.0 & 45.8 & 37.2 & 34.6 & 38.3 & 27.9 & 37.0 \\
Ovis2 34B & 34.0 & 50.0 & 41.1 & 37.0 & 40.6 & 36.2 & 33.3 & 31.8 & 38.7 \\
InternVL3 38B & 38.0 & 46.6 & 38.5 & 56.8 & 40.6 & 35.7 & 37.0 & 32.0 & 39.8 \\
InternVL3.5 38B$^{\textbf{R}}$ & 38.0 & 54.4 & 43.5 & 50.9 & 40.9 & 31.3 & 33.9 & 31.5 & 39.8 \\
\midrule
\multicolumn{10}{c}{\textit{Large Open-Weight Models (\textgreater 38B)}} \\
\midrule
LLaVA OV 72B & 72.0 & 35.4 & 30.5 & 28.1 & 32.3 & 28.4 & 31.5 & 26.0 & 30.5 \\
Qwen 2.5 VL 72B & 72.0 & 49.7 & 37.2 & 32.8 & 44.0 & 33.3 & 35.4 & 25.3 & 37.1 \\
InternVL3 78B & 78.0 & 49.5 & 39.3 & 45.3 & 39.3 & 33.9 & 35.9 & 30.5 & 38.6 \\
GLM 4.5V 106B$^{\textbf{R}}$ & 106.0 & 51.8 & 43.2 & 52.1 & 45.8 & 35.9 & 40.9 & 33.1 & 42.6 \\
\midrule
\multicolumn{10}{c}{\textit{Specialized High-Resolution Models}} \\
\midrule
InternLM XC 2.5 7B & 7.0 & 28.4 & 25.3 & 27.6 & 29.9 & 27.1 & 29.9 & 28.6 & 28.2 \\
VILA HD 4K 8B & 8.0 & 31.0 & 30.7 & 25.5 & 30.2 & 29.2 & 28.6 & 28.4 & 29.4 \\
\midrule
\multicolumn{10}{c}{\textit{Proprietary Models}} \\
\midrule
GPT-5 (minimal)$^{\textbf{R}}$ & — & 53.6 & 43.5 & 63.0 & 47.1 & 36.5 & 40.9 & 32.8 & 44.0 \\
Gemini 2.5 Pro$^{\textbf{R}}$ & — & 65.9 & 61.2 & 66.7 & 54.7 & 51.8 & 39.8 & 36.7 & 52.8 \\
GPT-5 (high)$^{\textbf{R}}$ & — & 63.8 & 54.4 & 70.3 & 58.1 & 51.0 & 46.9 & 41.1 & 53.9 \\
\bottomrule
\end{tabular}
\vspace{-0.3cm}
}
\end{table*}

%% file: iclr2026/tables/user_study_table.tex
\begin{table}[ht]
\centering
\caption{User Study Results. For each context scope, we report Accuracy (\%) for both Natural Language (NL) and JSON answer options. Human performance with NL is shown for reference.}
\label{table:user-study}
\small
\resizebox{0.9\linewidth}{!}{
\begin{tabular}{@{}lcccccc@{}}
\toprule
\multirow{2}*{\textbf{Model}}  &
\multicolumn{2}{c}{\textbf{Without Context}} &
\multicolumn{2}{c}{\textbf{Focused Context}} &
\multicolumn{2}{c}{\textbf{Whole Document Context}} \\
\cmidrule(lr){2-3} \cmidrule(lr){4-5} \cmidrule(lr){6-7}
&
\textbf{NL} &
\textbf{JSON} &
\textbf{NL} &
\textbf{JSON} &
\textbf{NL} &
\textbf{JSON} \\
\midrule
\textit{Human} & 27.5 & \textemdash & 77.5 & \textemdash & 65.0 & \textemdash \\
\midrule
InternVL3.5 8B$^{\textbf{R}}$ & 49.3 & 28.4 & 76.3 & 47.4 & 56.8 & 35.1 \\
InternVL3.5 38B$^{\textbf{R}}$ & 53.7 & 25.3 & 76.3 & 71.1 & 70.3 & 40.5 \\
Qwen 2.5 VL 72B                & 47.8 & 38.8 & 65.8 & 65.8 & 43.2 & 48.6 \\
Gemini 2.5 Pro$^{\textbf{R}}$ & 70.1 & 37.3 & 81.6 & 65.8 & 83.8 & 37.8 \\
\bottomrule
\end{tabular}}
\vspace{-0.3cm}
\end{table}

%% file: iclr2026/tables/no_context_ablation_table.tex
\begin{table}
\centering
\caption{Impact of answer representation on without-context performance and visual reliance. Accuracy is reported for the \emph{Ident} task. $R$ is computed according to Eq.~\ref{eq:vis-reliance}. using \textit{Focused} context as the with-context baseline.}
\label{table:nocontext-ablation}
\scriptsize
\resizebox{0.6\linewidth}{!}{%
\begin{tabular}{@{}l|cc|cc@{}}
\toprule
\multirow{2}*{\textbf{Model}} & \multicolumn{2}{c|}{\textbf{Natural Language}} & \multicolumn{2}{c}{\textbf{JSON}} \\
\cmidrule(lr){2-3} \cmidrule(lr){4-5}
 & \textbf{Accuracy} & \textbf{$R$} & \textbf{Accuracy} & \textbf{$R$} \\
\midrule
InternVL3.5 8B  & 45.6 & 17.1 & 28.6 & 29.3 \\
InternVL3.5 38B & 52.9 & 22.5 & 26.3 & 38.1 \\
Qwen 2.5 VL 72B  & 49.7 & 16.1 & 36.5 & 20.8 \\
Gemini 2.5 Pro   & 61.2 &  43.8 & 37.8 & 45.2 \\
\bottomrule
\end{tabular}
}
\end{table}

%% file: iclr2026/sections/conclusions.tex
\section{Conclusions}
We introduce \method, a multimodal benchmark for evaluating LMMs on real-world scientific inconsistencies. We show that even top-performing models struggle with cross-modal reasoning and long-context grounding, while structured answer formats mitigate linguistic shortcuts. This work highlights limitations of LMMS as scientific assistants and motivates future improvements in filtering pipelines, cross-domain datasets, and debiasing strategies for long-form MCQs evaluation.

\myparagraph{Limitations.} 
% We acknowledge that our benchmark currently focuses on AI-domain papers from ICLR 2025 and draws only from rejected submissions to capture unresolved errors. Future work could expand to include more diverse research fields and to analyze the potentially more subtle inconsistencies that persist in accepted papers.
Our benchmark is limited in scope: it currently focuses on AI-domain papers from ICLR 2024 \& 2025 and emphasizes rejected submissions to capture unresolved errors. As a result, both the scale and domain coverage are restricted. Future work should expand to other fields and venues, and explore inconsistencies that may persist in accepted papers, offering a broader and more representative testbed.

% \subsection{Future Work}

% \begin{itemize}
%     \item Better LLM filtering pipeline for OpenReview data
%     \item More data from different conferences from different fields
%     \item \textbf{Debiasing technique for long MCQ in natural language}
% \end{itemize}

%% file: iclr2026/sections/ethics.tex
\section{Ethics Statement}

% \subsection{Compliance of Copyright}

% \wei{ethics statement should be at the end of the main text before reference. It does not count toward the page limit, but should not be more than 1 page. }
% \\\wei{compliance of copyright ...  mention licence of Openreview papers}
This work introduces PRISMM-Bench, a benchmark for evaluating multimodal large language models (MLLMs) on scientific document understanding. In developing the benchmark, we exclusively use publicly available research papers from ICLR 2024 \& 2025, which are distributed under the Creative Commons Attribution 4.0 (CC-BY 4.0) license. This license explicitly permits redistribution, remixing, and adaptation of the material with proper attribution, and we ensure that all source materials are used in full compliance with these terms.

Our study also includes a small-scale user study to establish a human baseline. All participants were experienced researchers, voluntarily consented to take part, and no personally identifying information was collected or reported. The study design posed no foreseeable risks to participants and did not involve vulnerable populations.

We recognize that benchmarks can influence the direction of future model development. While PRISMM-Bench may highlight weaknesses in existing systems, it is not intended to facilitate misuse, such as adversarial attacks on models, but rather to promote more robust and trustworthy scientific document analysis. We release the benchmark with the goal of supporting transparent, reproducible, and ethical research, in line with the ICLR Code of Ethics.

No conflicts of interest, sensitive data, or privacy concerns arise in this work.

% All papers submitted to the International Conference on Learning Representations (ICLR) 2025 adopted the Creative Commons Attribution 4.0 (CC-BY 4.0) license. This license allows others to distribute, remix, adapt, and build upon the work, even for commercial purposes, as long as they provide appropriate credit to the original author(s). Therefore, all ICLR 2025 papers used in this paper are utilized in accordance with this license's terms, ensuring proper use and attribution of the source material.

% The icons used in \autoref{fig:pipeline} are sourced from Flaticon\footnote{https://www.flaticon.com} by the creators \textit{LAB Design Studio, catkure, Muhammad Ali and Uniconlabs}.

%% file: iclr2026/sections/appendix.tex
\appendix
\section*{Appendix}
% This is an appendix. \wei{appendix has to be after references}

% In the appendix, we first discuss the \textbf{source code} (App.~\ref {app:source_code}) and \textbf{qualitative examples} (App.~\ref{app:dataset-viewer}), including a dataset viewer and examples of text-table and figure-equation inconsistencies. 

In the appendix, we first discuss illustrative \textbf{qualitative examples} (App.~\ref{app:dataset-viewer}), including examples of text-table and figure-equation inconsistencies. We then provide a comprehensive \textbf{list of assets} (App.~\ref{app:assets}), detailing the data sources, licenses, and models used, including both open-source and proprietary ones. The \textbf{ablations section} (App.~\ref{app:ablations}) explores the impact of rasterization resolution on model performance, extends the analysis of the effectiveness of Chain-of-Thought (CoT) reasoning with a detailed case study, and validates our choice of the multiple-choice question format against open-ended evaluation. The \textbf{dataset construction section} (App.~\ref{app:data-coll}) explains our refined methodology for sourcing, filtering, and annotating inconsistencies from scientific papers, including a discussion of the annotation criteria, a custom-built annotation tool, and dataset statistics. Next, we detail the \textbf{process of generating LLM-based questions} (App.~\ref{app:task-generation}) for our benchmark tasks (Inconsistency Identification, Remedy, and Pair-Match), including our debiasing strategies using structured JSON representations. Finally, we provide details on the \textbf{user study implementation and representativeness} (App.~\ref{app:user-study}), the \textbf{full LLM prompts} (App.~\ref{app:prompts}) used for various tasks, and \textbf{screenshots} (App.~\ref{app:ann-app} and \ref{app:survey-app}) of our annotation and survey applications to provide a clear understanding of our methodology.

% \section{Source Code}\label{app:source_code}
% We provide the full source code for (1) the creation and evaluation of the benchmark dataset, (2) the annotation viewer and (3) the survey web app in the supplementary materials. More information on the structure and instructions on how to run the code is provided in a \texttt{`README.md`} file in \texttt{`source\_code\/`} folder of the material. Upon acceptance, we will release the source code publicly.

\section{Qualitative Examples}
\label{app:dataset-viewer}
% Alongside the source code, the supplementary materials contain an annotation viewer, which can be used to visually explore out dataset in the web browser. 
% The viewer can be used by opening the \texttt{`index.html`} file within the \texttt{`annotation\_viewer\/`} folder in a modern web browser (suggested Google Chrome).
% The viewer can be launched by opening the \texttt{index.html} file inside the \texttt{annotation\_viewer\/} folder (we recommend using Google Chrome).

% \wei{we can also show more qualitative examples of the tasks, }\\
We show qualitative examples of a text–table inconsistency in Fig.~\ref{fig:qualitative_example_table_text_inconsistency} and a figure-equation inconsistency in Fig.~\ref{fig:qualitative_example_figure_equation_inconsistency}, together with their corresponding evaluation tasks.

%Again, the full dataset will be released publicly upon acceptance.

\input{iclr2026/fig_tex/qualitative_example_table_text_inconsistency}

\input{iclr2026/fig_tex/qualitative_example_figure_equation_inconsistency}

% \section{Clarification of Usage of LLM for Paper Writing}
% \wei{This clarification is required by ICLR 2026.}

\section{List of Assets}\label{app:assets}

% \wei{very important!! clarification of the license of the ICLR paper}

% \wei{list of the data sources, implementation sources, model weights,  proprietary model version etc. }

Our images and annotations are sourced from publicly available datasets, and we distribute our data in compliance with the licensing terms of the original sources.

The document and review data source can be found here:
\begin{itemize}
    \item ICLR 2024 on OpenReview (\href{https://openreview.net/group?id=ICLR.cc/2025/Conference}{https://openreview.net/group?id=ICLR.cc/2024/Conference}): All papers were released under the CC BY 4.0 license.
    \item ICLR 2025 on OpenReview (\href{https://openreview.net/group?id=ICLR.cc/2025/Conference}{https://openreview.net/group?id=ICLR.cc/2025/Conference}): All papers were released under the CC BY 4.0 license.
\end{itemize}

The list of source code and model weights can be found here:
\begin{itemize}
    \item Qwen2.5-VL (\href{https://github.com/QwenLM/Qwen2.5-VL}{https://github.com/QwenLM/Qwen2.5-VL}): Released under the Apache-2.0 license.
    \item LLaVA-NeXT (\href{https://github.com/LLaVA-VL/LLaVA-NeXT}{https://github.com/LLaVA-VL/LLaVA-NeXT}): Released under the Apache-2.0 license.
    \item Gemma 3 (\href{https://github.com/google-deepmind/gemma}{https://github.com/google-deepmind/gemma}): Released under the Apache-2.0 license.
    \item Ovis 2 (\href{https://github.com/AIDC-AI/Ovis}{https://github.com/AIDC-AI/Ovis}): Released under the Apache-2.0 license.
    \item InternVL (\href{https://github.com/OpenGVLab/InternVL}{https://github.com/OpenGVLab/InternVL}): Released under the MIT license.
    \item GLM-V (\href{https://github.com/zai-org/GLM-V}{https://github.com/zai-org/GLM-V}): Released under the Apache-2.0 license.
    \item Mistral NeMo (\href{https://github.com/mistralai/mistral-inference}{https://github.com/mistralai/mistral-inference}: Released under Apache-2.0 license
    \item vLLM (\href{https://github.com/vllm-project/vllm}{https://github.com/vllm-project/vllm}): Released under the Apache-2.0 license.
    %\item LMDeploy (\href{https://github.com/InternLM/lmdeploy}{https://github.com/InternLM/lmdeploy}): %Released under the Apache-2.0 license.
    \item MinerU (\href{https://github.com/opendatalab/MinerU}{https://github.com/opendatalab/MinerU}): Released under the AGPL-3.0 license.
\end{itemize}

The list of proprietary models used can be found here:
\begin{itemize}
    \item Google Gemini (\href{https://deepmind.google/models/gemini/flash/}{https://deepmind.google/models/gemini/flash/}): Used in version Gemini Flash 2.5 and Gemini Pro 2.5, released on June 17, 2025.
    \item OpenAI GPT (\href{https://github.com/LLaVA-VL/LLaVA-NeXT}{https://github.com/LLaVA-VL/LLaVA-NeXT}): Used in version GPT-5, released on August 7, 2025.
\end{itemize}

\section{Ablations}\label{app:ablations}

\subsection{Impact of Rasterization Resolution}\label{app:dpi}
Scientific papers contain dense text and fine-grained visual elements such as axis labels, annotations, and subscripts, which are often crucial for detecting subtle inconsistencies. To test whether rasterization resolution impacts detection performance, we varied the DPI used to extract images from the PDF and evaluated a representative set of strongest proprietary and open-weight models of different sizes on the \textit{Inconsistency Ident} task with \textit{Focused Context}, keeping all other settings fixed.

\input{iclr2026/tables/dpi_ablation_table}

\myparagraph{Low resolutions harm performance.} Most models showed significant drops at 72 DPI, up to -20.1\% for InternVL3.5 38B. Open-weight models were generally more vulnerable, though even Gemini 2.5 Pro declined by -3.3\%. Surprisingly, VILA HD, despite being trained for high-resolution inputs, showed a slight accuracy gain at this lower setting.
% The one counterintuitive exception is VILA HD which - despite being specifically trained for high-resolution input - experienced a slight increase in identification accuracy on the lower resolution.

\myparagraph{Higher resolutions do not always improve accuracy.} Increasing from 144 to 300 DPI yielded mixed outcomes. While Gemini 2.5 Pro and Ovis2 34B benefited slightly, InternVL models performed worse, and VILA HD again failed to leverage the higher fidelity despite its specialized training. This suggests that additional detail can sometimes overwhelm global reasoning or misalign with training distributions. 
% from the increased level of detail, the improvement was comparatively minor compared to the drop in performance suffered by both InternVL models. Again, VILA HD failed to utilize the higher resolution, despite being explicitly trained on high DPI document images \citep{vila}. This suggests higher fidelity can be beneficial but may also harm the reasoning ability by dominating the global context or misaligning with training distributions, depending on the architecture \citep{llava-hd}.

\myparagraph{Resolution sensitivity is architecture-specific.} Overall, the assumption that higher resolution improves inconsistency detection does not hold universally. Performance varies with model design and pretraining data, and high-resolution training does not guarantee an edge in handling scientific inconsistencies. Careful DPI control is critical for fair evaluation. In our benchmark, 144 DPI provides a practical balance of visual clarity, computational cost, and cross-model comparability. 
% Instead, we observe a nuanced interaction between resolution, model architecture, and pretraining data. Training models on high resolution data does not automatically give them an advantage in detecting the subtle nuances found in scientific paper inconsistencies. During evaluation, it is therefore important to control DPI carefully, as overly low or high settings can unfairly penalize certain models. Our choice of 144 DPI provides a robust balance between detail, computational efficiency, and comparability across models.

\subsection{Case Study: Chain-of-Thought Reasoning}
\label{app:cot}

To illustrate the effect of chain-of-thought reasoning discussed in Section \ref{sec:cot}, we present a representative example in Fig.~\ref{fig:cot_figure}. The figure reports results for \textit{Unique Successful Jailbreaks}, a count-based metric that is strictly non-negative. However, some error bars extend below the zero line on the y-axis—an inconsistency that invalidates the figure.

\begin{center}
    \centering
    \includegraphics[width=1\linewidth]{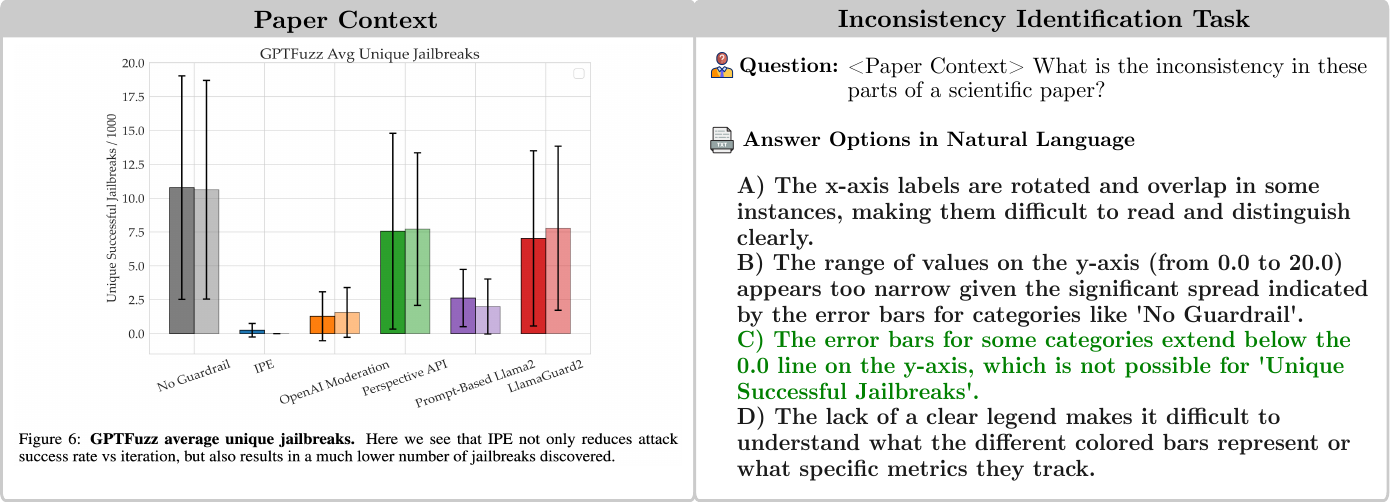}
\captionof{figure}{Inconsistency example for case study. Right: Visual context. Left: Question and answer options for \textit{Ident} task. Natural language options used for ease-of-comprehension, LMM was tasked using JSON.}\label{fig:cot_figure}
\end{center}

Without reasoning, InternVL3.5 38B selected the distractor “x-axis labels overlap,” (option A) justifying it with a generic but factually incorrect critique, as the labels were perfectly legible. The model defaulted to a template-like response rather than verifying claims against visual evidence. 

% Without reasoning, InternVL3.5 38B selected the distractor “x-axis labels overlap,” justifying its choice with the following explanation: 
% \begin{quote} 
% \textit{“The x-axis labels in the figure overlap, making them difficult to read. This violates the expectation that x-axis labels should be readable.”} 
% \end{quote}

% This rationale is linguistically coherent but factually incorrect, as the x-axis labels are clearly legible. The model defaulted to a generic figure-critique template rather than verifying the claim against the visual evidence.

In contrast, the reasoning-enabled model produced a systematic chain-of-thought: (1) ruling out label overlap (by observing the labels were \textit{'spaced out and readable'}), (2) confirming the y-axis range was sufficient (noting all data was \textit{'within the 0-20 range'}), (3) dismissing legend critiques (since a legend \textit{'isn't necessary'} when bars are directly labeled), and (4) crucially, identifying the logical error of error bars that \textit{'shouldn't go below zero if the metric [...] can't be negative'}). This stepwise elimination and domain-aware inference led to the correct answer. The full reasoning chain of InternVL3.5 38B is available in Fig.~\ref{fig:cot_chain}.
% that error bars extending below zero contradict the semantics of the metric, since counts cannot take negative values. This logical consistency check enabled the correct identification of the inconsistency.

This case highlights two key strengths of reasoning: (1) systematic elimination of distractors, and (2) integration of domain knowledge (e.g. non-negativity of counts) with visual grounding. Although reasoning increase output length (average of 473 tokens per query in our case), it substantially improves multimodal consistency and robustness, making CoT a key mechanism for handling subtle scientific document inconsistencies. 
% While this comes at the cost of longer outputs (more tokens per query), the improvement in multimodal grounding and structured error identification demonstrates that CoT is a crucial mechanism for handling the subtle inconsistencies found in scientific documents.

% \subsubsection{Full Reasoning Output for Chain-Of-Thought Analysis}
% \label{app:cot-ablation}

% We show the full reasoning output by InternVL3.5 38B on inconsistency \texttt{3MDmM0rMPQ} for the \textit{Ident} task in Fig.~\ref{fig:cot_chain}.

\input{iclr2026/fig_tex/cot_ablation_abstract}

\subsection{Validity of Multiple-Choice vs. Open-Ended Evaluation}
\label{app:mcq-vs-open-ended}
While open-ended generation is closer to real-world usage, the choice of a multiple-choice question (MCQ) format for our benchmark was motivated by the need for a reliable, reproducible, and bias-controlled evaluation protocol. To validate this choice, we conducted additional experiments comparing our MCQ format with an open-ended evaluation using an LLM-as-a-judge approach.

\myparagraph{Setup.} We tested a representative subset of the identification task using open-ended responses from eight models. Their outputs were rated by three different judges---Gemini 2.5 Pro, GPT-5 (high), and GLM 4.5V 106B---on a Likert-5 scale measuring semantic alignment with the ground-truth inconsistency. Each judge was evaluated both with and without \textit{Focused Context}.

\input{iclr2026/tables/mcq-vs-open-ablation}

\myparagraph{Analysis of Open-Ended Evaluation.} 
As shown in Table~\ref{tab:mcq-vs-open-ablation}, our open-ended evaluation revealed several critical limitations that undermine its suitability for a standardized benchmark. First, we observed significant \textbf{inter-judge inconsistency}, where LLM judges produced diverging scores and rankings for identical model outputs. For instance, Gemini 2.5 Pro and GPT-5 frequently disagreed on absolute scores, leading to rank shifts of two to three positions for certain models depending on the judge employed. Furthermore, GLM 4.5V 106B proved particularly unstable in its role as a judge, further complicating reproducibility.

A second challenge relates to \textbf{non-deterministic scoring} in proprietary reasoning models. Since a temperature of zero is not always available for models such as GPT-5, repeated evaluations of the same answer do not yield deterministic results. This introduces noise into the benchmarking process, which is especially problematic when performance differences between high-performing models are small. Finally, we identified a clear \textbf{judge-model similarity bias}, where judges tended to rate outputs from their own model families more favorably. This phenomenon, documented in prior evaluations such as MT Bench \citep{zheng2023judging} and AlpacaEval \citep{dubois2024alpacaeval}, makes it difficult to disentangle true model capability from stylistic alignment with the judge.

\myparagraph{Conclusion.} In contrast to open-ended evaluation, the MCQ format remains more reproducible, providing deterministic scoring, controlled difficulty through distractors, and identical evaluation conditions without dependence on third-party proprietary judges. Combined with our JSON-based debiasing strategies, the MCQ format provides a robust and reliable framework for \method.

\section{Dataset Construction}
% \wei{mention details of dataset construction that are not included in the main paper}\\

\subsection{Review Sourcing}\label{app:data-coll}

\myparagraph{Initial Exploration with Regex Matching.}
Before finalizing the review sourcing strategy described in the main paper, we conducted an exploratory study to detect potential inconsistencies mentioned in reviews for ICLR 2024 using a simple regular expression (regex) approach. Reviews were accessed via the OpenReview API\footnote{\url{https://docs.openreview.net/reference/api-v2}}, focusing on the \textit{“Weaknesses”} and \textit{“Questions”} sections, which were most likely to contain critical feedback. Each sentence was parsed for co-occurrence of terms related to inconsistencies (e.g., “mismatch", “conflict") and references to visual elements (e.g., “figure", “table", “equation"). The pseudocode for this procedure is shown below:

\begin{algorithm}
\caption{Pseudocode for regex matching}
\label{alg:regex-pseudocode}
\begin{algorithmic}[1]
% \State \Call{DefinePattern}{inconsistency\_pattern, 
%   \texttt{r'(inconsisten|mismatch|}\\ \texttt{doesn[']t match|not match|conflict|discrepanc)'} 
%   }
%     % \verb|r'(inconsisten|mismatch|\\doesn[']t match|not match|conflict|discrepanc)'|}
% \State \Call{DefinePattern}{visual\_pattern, 
%   \texttt{r'(figure|fig.?|table|graph|plot|}\\\texttt{image|diagram|equation)'}}
%     % \verb|r'(figure|fig.?|table|graph|plot|\\image|diagram|equation)'|}
\State \Call{DefinePattern}{inconsistency\_pattern,
  \texttt{r'(inconsisten | mismatch | doesn[']t match | not match |conflict | discrepanc)'}}
\State \Call{DefinePattern}{visual\_pattern,
  \texttt{r'(figure | fig.? | table | graph | plot | image | diagram | equation)'}}
\Statex
\State results $\gets$ [ ]
\For{each review in reviews}
    \State sections $\gets$ \Call{ExtractSections}{review}
    \For{each section in sections}
        \State sentences $\gets$ \Call{SplitIntoSentences}{section}
        \For{each sentence in sentences}
            \If{\Call{Matches}{sentence, inconsistency\_pattern} \textbf{and} \Call{Matches}{sentence, visual\_pattern}}
                \State \Call{Append}{results, sentence}
            \EndIf
        \EndFor
    \EndFor
\EndFor
\State \textbf{return} results
\end{algorithmic}
\end{algorithm}

Manual inspection confirmed that reviews indeed contained valuable references to visual-textual mismatches. However, two limitations emerged:  
(1) regex captured only strict keyword formulations, missing paraphrased or indirect mentions of inconsistencies, and  
(2) many inconsistencies referenced papers that had been updated after rebuttal, making it impossible to locate the original errors in the PDF versions available through OpenReview.

\myparagraph{Refined Collection Strategy.}
To address these issues, we refined our strategy in two ways:
\begin{enumerate}
    \item \textbf{Conference selection:} We shifted our focus on papers without author rebuttals. This ensured that flagged inconsistencies were more likely to remain in the available PDFs.  
    \item \textbf{LLM filtering:} Instead of regex, we employed \textit{Mistral Nemo 2407} at a low temperature to summarize reviews and extract candidate inconsistency statements. This approach captured non-strict formulations (e.g., “does not align with” instead of “mismatch”) and produced structured outputs, making them easier to present to annotators in the verification interface.
\end{enumerate}

This refinement reduced our initial pool of 120,329 ICLR reviews to 18,009 reviews. The LLM outputs were stored in structured JSON format, with each paper ID associated with a list of flagged inconsistencies.

\myparagraph{Example Output.}
Throughout the appendix, we are going to illustrate our data preparation pipeline use the paper ID \texttt{vXSCD3ToCS}\footnote{\url{https://openreview.net/forum?id=vXSCD3ToCS}} as an example. 
We illustrate the example out after the LLM-assisted filtering in Fig.~\ref{fig:example_output_after_llm_assisted_filtering}. 
% After the LLM filtering, the output was represented as:

\input{iclr2026/fig_tex/example_output_after_llm_assisted_filtering}

% \begin{verbatim}
% "vXSCD3ToCS": {
%   "has_inconsistency": true,
%   "inconsistencies": [
%     "Table 6: The performance improvement from using 20 years of data 
%      (MAE of 20.24) compared to 1 year (MAE of 21.90) in the 3-day 
%      setting is marginal, contradicting the emphasis on the scale of 
%      DynST as a major contribution.",
%     "Figure 2: The bottom left corner shows a road segment between two 
%      points that is not represented as an edge in the topology, which 
%      appears inconsistent with the actual road network.",
%     "Figure 5: The description of the figure suggests changes in sensors 
%      deployment, not actual road network dynamics, which contradicts 
%      the paper's claim of considering dynamic road network topology."
%   ]
% }
% \end{verbatim}

This structured representation provided a natural starting point for the subsequent manual verification stage described in App.~\ref{sub:ann-process}. % \wei{is this reference correct? Section G shows screenshots of annotation interface.}

% TODO

% \wei{details of the Next.js annotation tool}

% \wei{Details of manual annotation, which are human manually annotated? which are automatically generated? How are they automatically generated?
% the details are missing.   we can give some brief description in the main paper and refer to details in appendix.   Details of metadata.  The amount of time the manual annotation took. }

\subsection{Annotation Process}
\label{sub:ann-process}

\myparagraph{Annotator Background.} The annotation was conducted by the first author, who has an advanced background in Computer Science and Machine Learning. A consistent annotation standard was maintained throughout the project; any ambiguous or borderline cases were discussed with senior researchers until a consensus was reached.

\myparagraph{Annotation Criteria.}
During manual verification, the annotator judged each reviewer comment against the following criteria:
\begin{enumerate}
    \item The comment reflects objective feedback rather than a subjective suggestion.
    \item The comment describes an inconsistency involving two contradicting facts.
    \item Both conflicting parts can be located in the PDF.
    \item The inconsistency can be identified without deep domain-specific expertise (focus on visual/document-level inconsistencies).
    \item The inconsistency is significant and factual, not a minor typo or stylistic choice.
\end{enumerate}

\myparagraph{Annotation Interface.}
We implemented a custom web-based tool in \texttt{Next.js}. The interface displayed the reviewer’s comment (extracted by the LLM) alongside the corresponding paper embedded as a PDF viewer (compare Fig.~\ref{fig:ann-app-0}, Fig.~\ref{fig:ann-app-1} for screenshots of the app's interface). The annotator could:
\begin{itemize}
    \item Read the reviewer’s comment and decide whether it fulfilled criteria (1) and (2). If not, the instance was skipped.
    \item Search and inspect the relevant region of the embedded PDF.
    \item Toggle between \emph{one-part} and \emph{two-part} annotation modes:
    \begin{itemize}
        \item \textbf{One-part:} A single element (e.g., figure-caption inconsistency).
        \item \textbf{Two-part:} Two separate elements (e.g., figure vs. text, or figure vs. figure).
    \end{itemize}
    \item Specify for each part whether it was textual or visual:
    \begin{itemize}
        \item \emph{Visual:} Select the page, then draw a bounding box on a rendered thumbnail version the PDF page.
        \item \emph{Textual:} Enter the page and line number, and copy the relevant text snippet from the PDF.
    \end{itemize}
    \item Assign an inconsistency category via a drop-down menu.
    \item Provide a short free-text description of the inconsistency in their own words.
\end{itemize}

\myparagraph{Recorded Metadata.}
Each annotation combined automatically and manually collected fields:
\begin{itemize}
    \item \textbf{Automatically recorded:} element type (text or image), bounding box (relative coordinates) for visual selections, internal image identifier, reviewer’s original comment.
    \item \textbf{Manually entered:} page and line numbers, copied textual content, inconsistency category, and a short description by the annotator.
\end{itemize}

\myparagraph{Example Output.}
Annotations were stored in JSON format, combining visual/textual parts, reviewer comment, category, and description. We illustrate the example annotation output in JSON format in Fig.~\ref{fig:example_annot_output_in_json_format}. 

\input{iclr2026/fig_tex/example_annot_output_in_json_format}

% \begin{verbatim}
% {
%   "inconsistency_parts": [
%     {
%       "type": "image",
%       "page": 5,
%       "image_id": "vXSCD3ToCS_5_a1e8a4c6",
%       "bbox": { "x": 0.50, "y": 0.25,
%                 "width": 0.35, "height": 0.31 }
%     },
%     {
%       "type": "text",
%       "page": 5,
%       "line": 262,
%       "content": "The results demonstrate ..."
%     }
%   ],
%   "review_text": "Figure 2: The bottom left corner shows ...",
%   "category": "figure-text",
%   "description": "Missing edges between nodes compared to the claim."
% }
% \end{verbatim}

% \wei{dataset statistics, openreview paper length, number of pages, etc. }
% \myparagraph{Dataset Statistics.} 

\subsection{Statistics of Inconsistency Collection. } The annotation resulted in 384 inconsistencies from 353 ICLR papers. The average page count of each PDF was 16 pages. A total of 29 papers (7.6\%) had more than one inconsistency. The paper subjects were equally distributed across the range of topics for ICLR\footnote{\url{https://iclr.cc/Conferences/2025/CallForPapers}}, with (1) representation learning (26.0\%), (2) transfer learning (10.4\%), (3) generative models (9.6\%) and datasets and benchmarks (7.8\%) being the most frequent topics.

We identified 15 categories of inconsistencies based on the elements involved, with the distribution shown in Fig.~\ref{fig:type-dist}. The most common cases were figure–text mismatches and intra-figure inconsistencies.

% The manual verification produced a final dataset of 262 inconsistencies drawn from 242 ICLR papers. 

\input{iclr2026/fig_tex/distribution_inconsistency_types}

\subsection{LLM-Based Question Generation}
\label{app:task-generation}
\subsubsection{Inconsistency Identification (Ident)}

The \textit{Inconsistency Identification (Ident)} task was the first benchmark task we designed. For each annotated inconsistency, we instructed \textit{Gemini Flash} to generate a multiple-choice question (MCQ) with four options, one of which correctly describes the inconsistency.

\myparagraph{Inputs.}
As input to the model, we provided:
\begin{itemize}
    \item The annotated context (visual and/or textual parts).
    \item The annotator’s free-text description of the inconsistency.
\end{itemize}

\myparagraph{Prompt.}
After experimenting with several formulations, we found that a minimalist prompt yielded the most creative and plausible distractors. We provide the final version of the prompt in Fig.~\ref{fig:prompt_preparing_MCQ_inconsistency_identification}. 

\input{iclr2026/fig_tex/prompt_preparing_MCQ_inconsistency_identification}

% \begin{verbatim}
% You are a visual assistant that can analyze image and text excerpts from scientific papers. 
% You receive either one image, two images, or a pair of image and text that contain a visual 
% inconsistency flagged by reviewers. Alongside the content, you also receive a description 
% of the inconsistency. Based on these, generate a multiple-choice question testing the model's 
% ability to detect the inconsistency. Follow these strict rules:

% - The question should directly reference the provided content of the paper.
% - There must be exactly 4 answer choices.
% - Only one answer should correctly describe the inconsistency.
% - The 3 distractors must be plausible but incorrect. They should either be incorrect due 
%   to omission or subtle misinterpretations of the content.
% - Do not invent details beyond what is provided.
% - Clearly label the correct answer.
% \end{verbatim}

\myparagraph{Output Format.}
The model produced a structured output containing the question, the correct answer, and three distractor answers.

\myparagraph{Manual Verification.}
Each generated question underwent manual verification:
\begin{itemize}
    \item \textbf{Correct answer:} must (1) faithfully reflect the annotator’s description and (2) directly connect to the annotated context.
    \item \textbf{Distractors:} must (1) be grounded in the annotated inconsistency parts, (2) avoid obvious contradictions within the answer itself, (3) only mention elements present in the provided context, and (4) describe an inconsistency rather than confirming a correct fact from the paper
\end{itemize}
We also post-processed the text to remove stylistic artifacts often appended by the LLM, e.g. the parenthetical \textit{“, indicating an inconsistency.”}

\myparagraph{Example Output.} We illustrate an example of the generated multiple-choice question for inconsistency identification task in Fig.~\ref{fig:example_generated_MCQ_inconsistency_identification}. 

\input{iclr2026/fig_tex/example_generated_MCQ_inconsistency_identification}

% \begin{verbatim}
% {
%   "mcq": {
%     "default": {
%       "question": "What inconsistency is observed between Figure 2 and the accompanying text regarding the generated road network?",
%       "correct": "The visualization in Figure 2 shows missing edges between nodes, which contradicts the text's claim that the generated network perfectly matches the actual road network structure.",
%       "incorrect": [
%         "The visualization in Figure 2 shows extraneous edges between nodes, which contradicts the text's claim that the generated network perfectly matches the actual road network structure.",
%         "Figure 2 depicts only a disconnected portion of the road network, implying the algorithm did not generate the complete structure.",
%         "The blue nodes in Figure 2 are unevenly distributed, making it difficult to determine the precise road paths."
%       ],
%       "letters": [
%         "D",
%         "A",
%         "B",
%         "C"
%       ]
%     }
%   }
% }
% \end{verbatim}

\subsubsection{Debiasing the Inconsistency Identification Task}\label{app:debiasing_inconsistency_identification}

\myparagraph{Initial Observations.}
When first evaluating the \textit{Ident} task with \textit{Gemini 2.5 Flash}, we observed unexpectedly high accuracy:
\begin{itemize}
    \item $84.4\%$ with the original LLM-generated questions. (e.g.: \textit{``What inconsistency is observed between Figure 2 and the accompanying text regarding the generated road network?"})
    \item $79.4\%$ after replacing the LLM-generated question with the generic formulation:
    \textit{``What is the inconsistency in these parts of a scientific paper?"}
\end{itemize}
Even in a sanity check where the model was shown only the question and answer options (without the annotated context), performance remained at $57.6\%$ accuracy, far above the random baseline of $25\%$. This indicated strong reliance on linguistic cues in the answer phrasing.

\myparagraph{Mitigation Strategies.}
Moving forward, we solely employed the generic question formulation throughout all inconsistencies. For reducing the without context accuracy $Acc_{nc}$, we systematically explored ways of reducing linguistic priors by rewriting the answer options:
\begin{itemize}
    \item Normalizing answer length: $Acc_{nc}=48.1\%$.
    \item Filtering for MCQs where the correct answer is shortest: $Acc_{nc}=46.2\%$.
    \item Rephrasing distractors according to best practices in MCQ test design~\citep{mcq_design_review}: $Acc_{nc}=41.6\%$.
        \item Shortening all answer options into nominal style: $Acc_{nc}=38.2\%$.
\end{itemize}

While these interventions reduced bias, they did not remove it completely.

\myparagraph{Structured Representation: Evidence–Claim JSON.}
As a more robust solution, we abandoned free-form natural language and introduced a structured, human-readable JSON representation that removes stylistic cues while preserving the semantic contradiction. The schema is:

\begin{verbatim}
{
  "letter": "A" | "B" | "C" | "D",
  "attribute": str,
  "claim": {
    "source": "expectation" | str,
    "statement": str
  },
  "evidence": {
    "source": str,
    "statement": str
  }
}
\end{verbatim}

\myparagraph{Patterns.}  
Two patterns of contradiction are covered:
\begin{itemize}
    \item \textbf{Claim vs. Evidence:} A claim from one paper element is contradicted by evidence from another.
    \item \textbf{Expectation vs. Evidence:} A claim contradicts common expectations of scientific correctness. In this case, the claim's source is always "expectation"
\end{itemize}

We prompted \textit{Gemini 2.5 Flash} to convert the natural language MCQs into this structured format. The full prompt can be inspected in App.~\ref{app:prompt-evidence-claim}. A 20\% subset was manually validated for consistency.

\myparagraph{Effect on Model Behavior.}
This representation further reduced the no-context accuracy to $34.0\%$. Given the full context, accuracy on the new JSON format decreased from $79.4\%$ to $69.5\%$. However, the fraction of performance attributable to visual grounding (Eq. \ref{eq:vis-reliance}) increased from $51.4\%$ to $53.8\%$. Thus, the structured format acts as a regularizer, forcing models to rely more strongly on the provided paper context.

\myparagraph{Example Output.}
For the running example, we illustrate the example of debiased output in the evidence-claim JSON format for the inconsistency identification task in Fig.~\ref{fig:example_debiased_output_in_evidence_claim_json_format}. 

\input{iclr2026/fig_tex/example_debiased_output_in_evidence_claim_json_format}

% \begin{verbatim}
% {
%   "mcq": {
%     "default": {
%       "question": "What is the inconsistency in these parts of a scientific paper?",
%       "correct": {
%         "letter": "A",
%         "attribute": "edges",
%         "claim": {
%           "source": "text",
%           "statement": "perfectly matches"
%         },
%         "evidence": {
%           "source": "Figure 2",
%           "statement": "missing edges"
%         }
%       },
%       "incorrect": [
%         {
%           "letter": "C",
%           "attribute": "edges",
%           "claim": {
%             "source": "text",
%             "statement": "perfectly matches"
%           },
%           "evidence": {
%             "source": "Figure 2",
%             "statement": "extraneous edges"
%           }
%         },
%         {
%           "letter": "D",
%           "attribute": "network",
%           "claim": {
%             "source": "expectation",
%             "statement": "complete structure"
%           },
%           "evidence": {
%             "source": "Figure 2",
%             "statement": "disconnected portion"
%           }
%         },
%         {
%           "letter": "B",
%           "attribute": "nodes",
%           "claim": {
%             "source": "expectation",
%             "statement": "evenly distributed"
%           },
%           "evidence": {
%             "source": "Figure 2",
%             "statement": "unevenly distributed"
%           }
%         }
%       ],
%       "letters": [
%         "A",
%         "C",
%         "D",
%         "B"
%       ]
%     }
%   }
% }
% \end{verbatim}

\subsubsection{Inconsistency Remedy Task}

\myparagraph{Task Design.}
The \textit{Inconsistency Remedy (Remedy)} task extends beyond identifying an inconsistency to determining how it can be resolved. To avoid linguistic artifacts, we directly employed a structured representation in JSON format. This representation adapts the Evidence–Claim schema to a more action-oriented form, the \textbf{Target–Action JSON}:

\begin{verbatim}
{
  "letter": "A" | "B" | "C" | "D",
  "attribute": str,
  "target": str,
  "other_involved": str,
  "action": "modify" | "remove" | "add" | "reposition" | "replace",
  "edit_statement": str,
  "reason": str
}
\end{verbatim}

Here, attribute captures the element at issue, target specifies where the change is applied, other\_involved records additional parts if necessary, and the fields action, edit\_statement, and reason summarize the correction.

\myparagraph{LLM Conversion Process.}
We found that prompting an LLM to directly convert the natural language MCQs from the \textit{Ident} task into Target–Action JSON yielded the most reliable results in terms of readability and correctness. The prompt is depicted in Sec.~\ref{app:prompt-target-action}.

\myparagraph{Example Output.} 
Four our example used throughout this appendix, the task looks as follows:
We illustrate the example output in Target-Action JSON format for the inconsistency remedy task in Fig.~\ref{fig:example_output_target_action_json_format}. 

\input{iclr2026/fig_tex/example_output_in_target_action_json_format}

% \begin{verbatim}
% {
%   "mcq": {
%     "edit": {
%       "question": "What action needs to be taken to resolve the inconsistency in these parts of a scientific paper?",
%       "correct": {
%         "letter": "A",
%         "attribute": "edges",
%         "target": "figure_2",
%         "other_involved": "text",
%         "action": "add",
%         "edit_statement": "missing edges",
%         "reason": "contradicts claim"
%       },
%       "incorrect": [
%         {
%           "letter": "C",
%           "attribute": "edges",
%           "target": "figure_2",
%           "other_involved": "text",
%           "action": "remove",
%           "edit_statement": "extraneous edges",
%           "reason": "contradicts claim"
%         },
%         {
%           "letter": "D",
%           "attribute": "road network",
%           "target": "figure_2",
%           "other_involved": "algorithm",
%           "action": "modify",
%           "edit_statement": "disconnected portion",
%           "reason": "incomplete structure"
%         },
%         {
%           "letter": "B",
%           "attribute": "blue nodes",
%           "target": "figure_2",
%           "other_involved": null,
%           "action": "modify",
%           "edit_statement": "distribute nodes evenly",
%           "reason": "unclear paths"
%         }
%       ],
%       "letters": [
%         "A",
%         "C",
%         "D",
%         "B"
%       ]
%     }
%   }
% }
% \end{verbatim}

\subsubsection{Inconsistency Pair-Match Task}

\myparagraph{Task Design.}
The \textit{Inconsistency Pair-Match (Match)} task focuses on the subset of inconsistencies that involve two distinct visual parts. The model is presented with one element (text or visual) as the question context and must identify the corresponding inconsistent visual element among four options.

\myparagraph{Filtering of Eligible Cases.}
Not all inconsistency categories are suitable for pair matching. Categories where the contradiction is contained entirely within a single element (i.e., \emph{figure–caption}, \emph{figure-only}, \emph{table-only}, \emph{table–caption}, \emph{algorithm-only}) were excluded. This filtering left 135 out of the 262 inconsistencies in the dataset.

\myparagraph{Distractor Construction.}
To ensure challenging and fair distractors, we extracted all figures, tables, and equations from the 242 papers in our dataset using \textit{MinerU}\footnote{\url{https://github.com/opendatalab/MinerU}}, which produced image crops with unique IDs, modality labels, and page numbers. We then implemented a python script to sample distractors as follows:
%  \wei{also provide the link of tool in footnote}
\begin{itemize}
    \item Distractors were restricted to the same modality as the correct answer.
    \item Preference was given to elements appearing on the same page or on adjacent pages to the correct element, so that distractors were topically similar.
    \item Sampling was done within the same paper. Each paper contained enough visual elements of the same modality so we didn't have to fallback to using elements from other papers.
\end{itemize}
This procedure reduced trivial elimination strategies (e.g., selecting “the only figure among tables”) and forced models to consider fine-grained inconsistencies.

\myparagraph{Example Output.}
In the running example, the annotated inconsistency links an in-line text with a figure. The text is fixed as the question context, and the answer options are image IDs referring to extracted figures, with one image ID being the correct image cropped in the annotations. We show the example of output used in the inconsistency pair-match task in Fig.~\ref{fig:example_output_for_inconsistency_pair_match_task}. 

\input{iclr2026/fig_tex/example_output_for_inconsistency_pair_match_task}

% \begin{verbatim}
% {
%   "mcq": {
%     "part_pair": {
%       "question": "The results demonstrate that the adjacency matrix generated by our algorithm perfectly matches the actual road network structure.",
%       "correct": "vXSCD3ToCS_5_a1e8a4c6",
%       "incorrect": [
%         "vXSCD3ToCS_5_image_figure3",
%         "vXSCD3ToCS_5_image_figure4",
%         "vXSCD3ToCS_6_image_figure5"
%       ],
%       "letters": [
%         "D",
%         "A",
%         "C",
%         "B"
%       ]
%     }
%   }
% }
% \end{verbatim}

\section{User Study Implementation \& Statistics}
\subsection{Implementation of the User Study}
\label{app:user-study}

\myparagraph{Setup.} We conducted the user study in online form, using a custom web app. The participants were greeted with an onboarding screen, where they entered the following information to assess their eligibility to be included in the user study: (1) email address, (2) academic field, (3) academic level and(4) AI exposure. Afterwards, they were shown instructions for the survey and an introduction into the question formats and different context modalities. For each participants, ten tasks were randomly sampled from our dataset. For the first five tasks, the participants were shown the \textit{Focused Context} with the exact cropped images and/or text passages from the paper. For the last five tasks, the participants were instructed to open a link to the original PDF and use the whole document to answer the question. In this case, they were provided with the visual element they should focus on in the paper. Screenshots of the user interface are provided in appendix~\ref{app:survey-app}.

Upon submission, the following datapoints were saved automatically: (1) The task ID, (2) The chosen answer by the participants, (3) whether the task was correctly answered with/without context, (4) whether the question was accompanied by \textit{Focused Context} or \textit{Full Document Context} and (5) the time it took the participants to answer each question.

\myparagraph{Statistics.} Our eight participants all have a background in either artificial intelligence, computer science or mathematics at an academic level of PhD or higher (7 PhD, 1 postdoc). All stated to exhibit advanced exposure levels to AI, which we defined as being comfortable with reading, interpreting and critically evaluating AI scientific literature. The median answer time for questions without provided visual context was 45s, with \textit{Focused Context} 145s and with \textit{Whole Document Context} 169s. In 65\% of the cases, participants changed their answers once provided with the context. In total, participants processed 80 inconsistencies.

\subsection{Representativeness of the User Study Subset}
\label{app:user-study-represent}
To ensure that the 40 samples used in the user study are representative of the full benchmark, we clarify that they were randomly drawn from the full dataset without any filtering or cherry-picking. As for some models, such as InternVL3.5 38B and Qwen2.5 VL 72B, accuracies on this subset were higher than their full-benchmark performance, we assessed whether this deviation was expected by running a 100-trial simulation. In each trial, we repeatedly sampled 40 random instances and computed the resulting scores. 

As shown in Table~\ref{tab:user-study-proof}, the simulation results indicate that the user-study subset happened to be in the upper quartile of model difficulty for these two models—a comparatively ``easier'' slice of the benchmark for them. Such variation is expected with small sample sizes and does not reflect systematic bias in the subset selection. Importantly, this strengthens our human-LMM comparison: even on a subset where LMMs performed better than their typical accuracy, humans still outperformed the models by a clear margin under both focused and whole-document conditions. If the subset had been closer to the mean model difficulty, the human-model gap would have been even larger. Thus, our user study results likely understate the core conclusion that humans substantially outperform current LMMs at detecting multimodal scientific inconsistencies.

\input{iclr2026/tables/user-study-proof}

% \section{Implementation Details}
% \wei{model hyperparameters etc. } 

\section{LLM Prompts}\label{app:prompts}
Here we provide the full prompts used to instruct the LLMs. 

\subsection{LLM Prompt for Review Filtering}
\label{app:prompt-review-filtering}
We provide the prompt for LLM-based review filtering in Fig.~\ref{fig:prompt_llm_based_review_filtering}. Given the reviewer's comment, the model is instructed in a chain-of-thought prompting manner to systematically analyze each desired characteristic of a visual inconsistency. Few-shot examples help clarify the output format.
% \wei{introduce a bit about the prompt.}

\input{iclr2026/fig_tex/prompt_llm_based_review_filtering}

\subsection{LLM Prompt for converting into Evidence-Claim Format}\label{app:prompt-evidence-claim}
We provide the prompt for LLM-assisted conversion of natural language answers (for the inconsistency identification task) into evidence-claim JSON format in Fig.~\ref{fig:prompt_converting_into_evidence_json_format}. The evidence-claim JSON format is used as answer options in the inconsistency identification task.
The structured JSON-based answer representation is for mitigating the language biases in multiple-choice evaluation. 

\input{iclr2026/fig_tex/prompt_conversion_into_evidence_claim_json_format}

\subsection{LLM Prompt for converting into Target-Action Format}
\label{app:prompt-target-action}
We provide the prompt for LLM-assisted conversion of natural language answers (for the inconsistency identification task) into target-action JSON format in Fig.~\ref{fig:prompt_converting_into_target_action_json_format}. Based on the question-answer pairs in inconsistency identification task, we generate question with answers for the inconsistency remedy task. 
The target-action JSON format is used as answer options in the inconsistency remedy task. 
% \wei{double check if this interpretation is correct... }

\input{iclr2026/fig_tex/prompt_conversion_into_target_action_json_format}

\section{Screenshots of the Annotation App}
\label{app:ann-app}

We show some examples of the interface of the annotation tool in Fig.~\ref{fig:ann-app-0} and Fig.~\ref{fig:ann-app-1}.

\input{iclr2026/fig_tex/annotation_app_vis_first_part}

\input{iclr2026/fig_tex/annotation_app_vis_second_part}

\section{Screenshots of the Survey App}
\label{app:survey-app}

We show some examples of the interface of the survey web interface in Fig.~\ref{fig:survey-app-0}, Fig.~\ref{fig:survey-app-1} and Fig.~\ref{fig:survey-app-2}.

\input{iclr2026/fig_tex/survey_app_vis_first_part}

\input{iclr2026/fig_tex/survey_app_vis_second_part}

\input{iclr2026/fig_tex/survey_app_vis_third_part}

%% file: iclr2026/fig_tex/qualitative_example_table_text_inconsistency.tex
\begin{figure}[h]
\begin{center}
\includegraphics[width=0.74\linewidth]{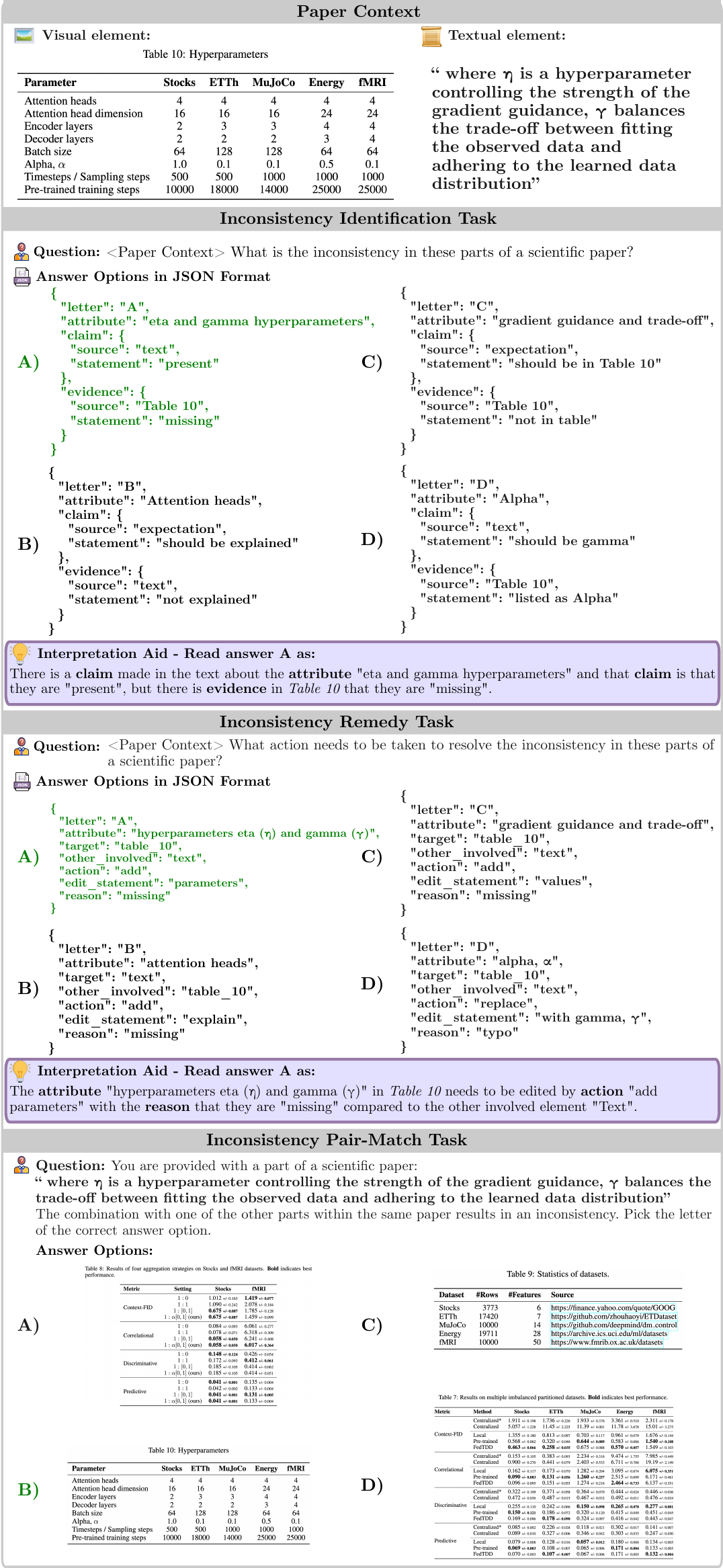}
\end{center}
\caption{A qualitative example of a text–table inconsistency and its corresponding evaluation tasks of \textit{Ident}, \textit{Remedy} and \textit{Match}.
}
\label{fig:qualitative_example_table_text_inconsistency}
\end{figure}

%% file: iclr2026/fig_tex/qualitative_example_figure_equation_inconsistency.tex
\begin{figure}[h]
\begin{center}
\includegraphics[width=0.85\linewidth]{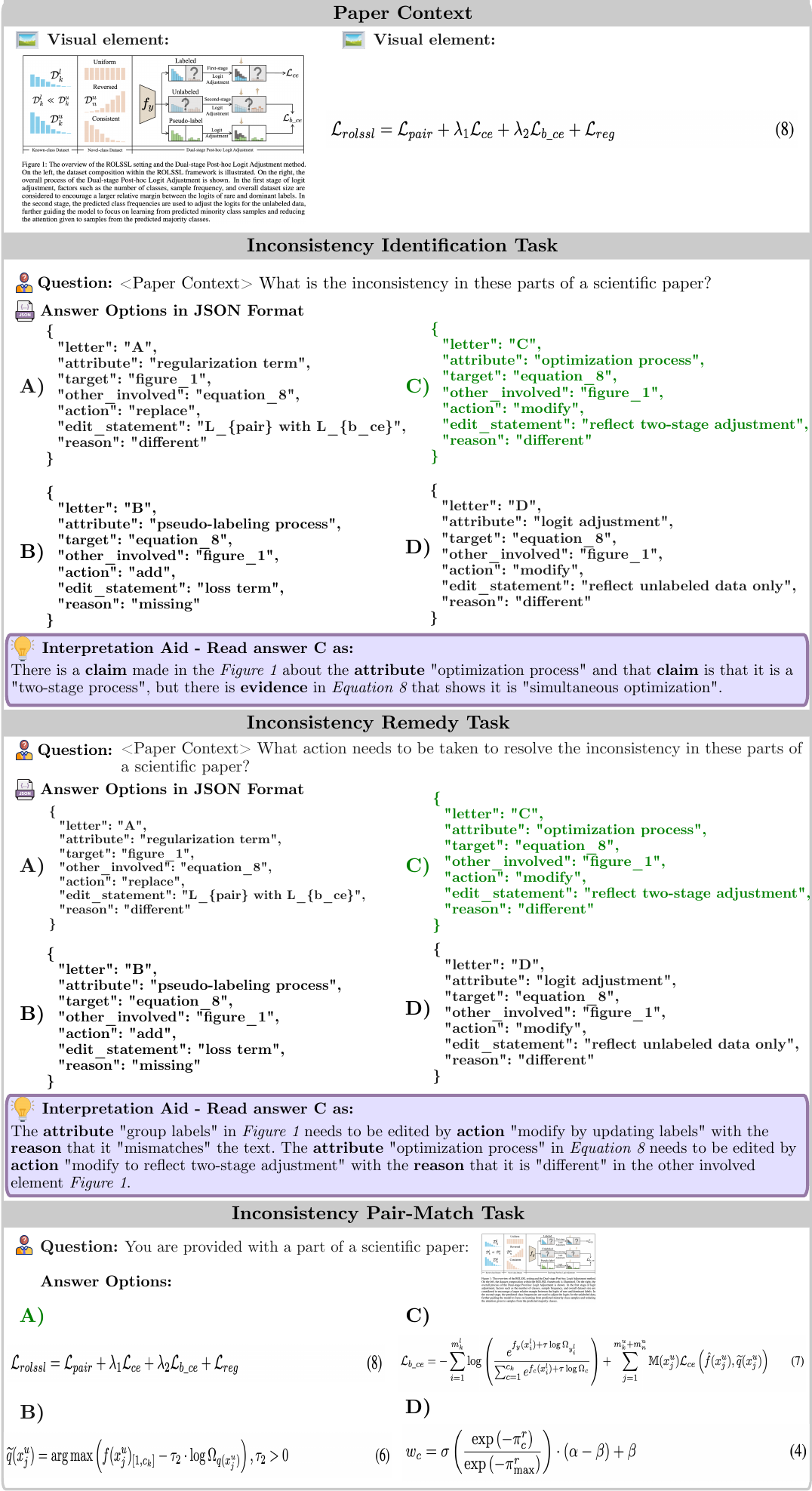}
\end{center}
\caption{A qualitative example of a figure-equation inconsistency and its corresponding evaluation tasks of \textit{Ident}, \textit{Remedy} and \textit{Match}.
}
\label{fig:qualitative_example_figure_equation_inconsistency}
\end{figure}

%% file: iclr2026/tables/dpi_ablation_table.tex
\begin{table}[ht]
\centering
\caption{
% Benchmark results for different DPI values. The accuracy is reported as a fraction of correct answers for the \emph{Ident} task with \emph{Focused Context}. Percentage change is calculated relative to the 144 DPI baseline.
Accuracy of LMMs under different rasterization resolutions. Results are reported for the \emph{Ident} task with \emph{Focused Context}. Percentage change is calculated relative to the 144 DPI baseline.
}
\label{table:dpi_benchmarks_inconsistency}
\small
\resizebox{\linewidth}{!}{%
\begin{tabular}{@{}l|cc|cc|cc|cc|cc@{}}
\toprule
\multirow{2}*{\textbf{DPI}}  & \multicolumn{2}{c|}{\textbf{VILA HD 4K 8B}} & \multicolumn{2}{c|}{\textbf{InternVL3.5 8B}} & \multicolumn{2}{c|}{\textbf{InternVL3.5 38B}} & \multicolumn{2}{c|}{\textbf{Ovis2 34B}} & \multicolumn{2}{c@{}}{\textbf{Gemini 2.5 Pro}} \\
\cmidrule(lr){2-3} \cmidrule(lr){4-5} \cmidrule(lr){6-7} \cmidrule(lr){8-9} \cmidrule(lr){10-11}
 & \textbf{Default} & \textbf{\% Change} & \textbf{Default} & \textbf{\% Change} & \textbf{Default} & \textbf{\% Change} & \textbf{Default} & \textbf{\% Change} & \textbf{Default} & \textbf{\% Change} \\
\midrule
72 & 31.3 & +1.3 & 42.7 & -12.7 & 45.8 & -20.1 & 43.1 & -13.1 & 67.2 & -3.3 \\
144 & 30.9 & -- & 48.9 & -- & 57.3 & -- & 49.6 & -- & 69.5 & -- \\
300 & 29.4 & -4.9 & 45.4 & -7.2 & 51.9 & -9.4 & 50.4 & +1.6 & 70.6 & +1.6 \\
\bottomrule
\end{tabular}
}
\end{table}

%% file: iclr2026/fig_tex/cot_ablation_abstract.tex
\begin{figure}[h]
    \centering
    \includegraphics[width=1\linewidth]{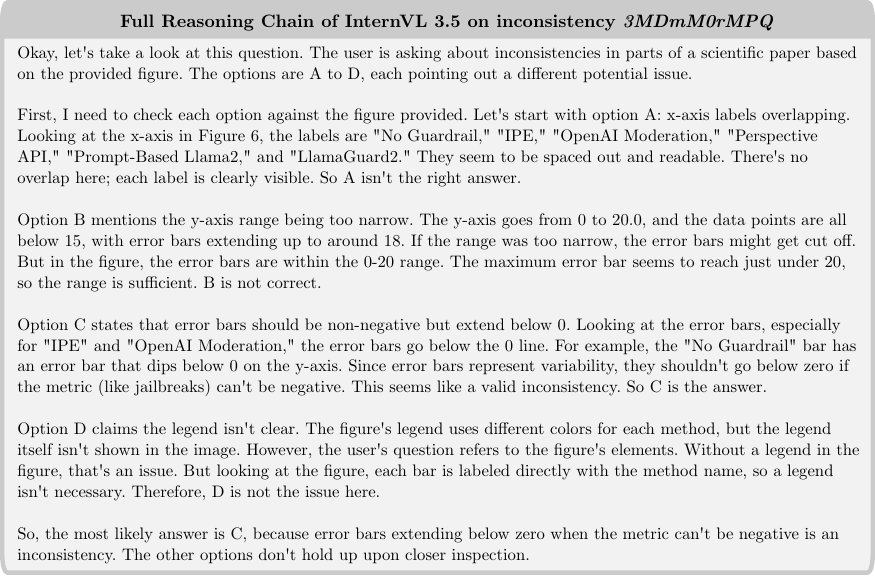}
    \caption{Raw reasoning output by InternVL3.5 38B on 3MDmM0rMPQ.}
    \label{fig:cot_chain}
\end{figure}

%% file: iclr2026/tables/mcq-vs-open-ablation.tex
\begin{table}[ht]
\centering
\caption{
LLM-as-a-judge evaluation. Scores are average Likert ratings (1-5); percentages indicate scores $>$ 3. R denotes the model's rank per judge.
}
\label{tab:mcq-vs-open-ablation}
\small
\resizebox{\linewidth}{!}{%
\begin{tabular}{@{}l|cc|cc|cc|cc|cc|cc@{}}
\toprule
\multirow{3}*{\textbf{Candidate Model}} & \multicolumn{4}{c|}{\textbf{Gemini 2.5 Pro}} & \multicolumn{4}{c|}{\textbf{GPT-5}} & \multicolumn{4}{c@{}}{\textbf{GLM 4.5V 106B}} \\
\cmidrule(lr){2-5} \cmidrule(lr){6-9} \cmidrule(l){10-13}
& \multicolumn{2}{c|}{\textbf{w/o context}} & \multicolumn{2}{c|}{\textbf{context}} & \multicolumn{2}{c|}{\textbf{w/o context}} & \multicolumn{2}{c|}{\textbf{context}} & \multicolumn{2}{c|}{\textbf{w/o context}} & \multicolumn{2}{c@{}}{\textbf{context}} \\
\cmidrule(lr){2-3} \cmidrule(lr){4-5} \cmidrule(lr){6-7} \cmidrule(lr){8-9} \cmidrule(lr){10-11} \cmidrule(lr){12-13}
 & \textbf{Score} & \textbf{R} & \textbf{Score} & \textbf{R} & \textbf{Score} & \textbf{R} & \textbf{Score} & \textbf{R} & \textbf{Score} & \textbf{R} & \textbf{Score} & \textbf{R} \\
\midrule
\textbf{Gemini Pro 2.5$\textsuperscript{R}$} & 2.80 (44.0\%) & \cellcolor{jkuLightGreen!50}2 & 3.06 (54.0\%) & \cellcolor{jkuGreen!50}\textbf{1} & 2.71 (40.0\%) & \cellcolor{jkuLightGreen!50}2 & 2.80 (44.0\%) & \cellcolor{jkuLightGreen!50}2 & 3.06 (54.0\%) & \cellcolor{jkuGreen!50}\textbf{1} & 3.50 (62.0\%) & \cellcolor{jkuGreen!50}\textbf{1} \\
\textbf{GPT-5 (high)$\textsuperscript{R}$} & 3.00 (54.0\%) & \cellcolor{jkuGreen!50}\textbf{1} & 2.70 (42.0\%) & \cellcolor{jkuLightGreen!50}2 & 2.76 (52.5\%) & \cellcolor{jkuGreen!50}\textbf{1} & 2.94 (52.0\%) & \cellcolor{jkuGreen!50}\textbf{1} & 3.02 (51.0\%) & \cellcolor{jkuLightGreen!50}2 & 3.32 (58.0\%) & \cellcolor{jkuLightGreen!50}2 \\
\textbf{InternVL 3.5 38B$\textsuperscript{R}$} & 2.22 (34.0\%) & \cellcolor{jkuPurple!30}6 & 2.26 (34.0\%) & \cellcolor{jkuCyan!40}4 & 2.06 (25.0\%) & \cellcolor{jkuCyan!40}4 & 1.98 (24.4\%) & \cellcolor{jkuCyan!40}4 & 2.68 (48.0\%) & \cellcolor{jkuYellow!50}3 & 2.84 (46.0\%) & \cellcolor{jkuYellow!50}3 \\
\textbf{Qwen 2.5 VL 72B} & 2.46 (38.0\%) & \cellcolor{jkuYellow!50}3 & 2.40 (38.0\%) & \cellcolor{jkuYellow!50}3 & 2.30 (28.3\%) & \cellcolor{jkuYellow!50}3 & 2.13 (26.7\%) & \cellcolor{jkuYellow!50}3 & 2.42 (34.0\%) & \cellcolor{jkuPurple!30}6 & 2.60 (40.0\%) & \cellcolor{jkuPurple!30}6 \\
\textbf{GLM 4.5V 106B$\textsuperscript{R}$} & 2.40 (40.0\%) & \cellcolor{jkuCyan!40}4 & 2.08 (28.0\%) & \cellcolor{jkuBlue!30}5 & 2.04 (22.0\%) & \cellcolor{jkuBlue!30}5 & 1.94 (22.0\%) & \cellcolor{jkuBlue!30}5 & 2.56 (42.0\%) & \cellcolor{jkuBlue!30}5 & 2.83 (47.9\%) & \cellcolor{jkuCyan!40}4 \\
\textbf{InternVL 3.5 8B$\textsuperscript{R}$} & 2.04 (28.0\%) & \cellcolor{jkuRed!30}7 & 1.94 (28.0\%) & \cellcolor{jkuPurple!30}6 & 1.94 (18.0\%) & \cellcolor{jkuRed!30}7 & 1.66 (18.0\%) & \cellcolor{jkuGrey!30}8 & 2.60 (42.0\%) & \cellcolor{jkuCyan!40}4 & 2.78 (42.0\%) & \cellcolor{jkuBlue!30}5 \\
\textbf{Gemma 3 12B} & 2.24 (34.0\%) & \cellcolor{jkuBlue!30}5 & 1.78 (24.0\%) & \cellcolor{jkuGrey!30}8 & 2.03 (19.3\%) & \cellcolor{jkuPurple!30}6 & 1.78 (18.4\%) & \cellcolor{jkuPurple!30}6 & 2.22 (32.0\%) & \cellcolor{jkuRed!30}7 & 2.50 (36.0\%) & \cellcolor{jkuRed!30}7 \\
\textbf{Ovis 2 34B} & 1.96 (20.0\%) & \cellcolor{jkuGrey!30}8 & 1.86 (22.0\%) & \cellcolor{jkuRed!30}7 & 1.83 (14.0\%) & \cellcolor{jkuGrey!30}8 & 1.74 (16.0\%) & \cellcolor{jkuRed!30}7 & 2.18 (32.0\%) & \cellcolor{jkuGrey!30}8 & 2.30 (32.0\%) & \cellcolor{jkuGrey!30}8 \\
\bottomrule
\end{tabular}
}
\end{table}

%% file: iclr2026/fig_tex/example_output_after_llm_assisted_filtering.tex
\begin{figure}[h]
    \centering
    \includegraphics[width=1\linewidth]{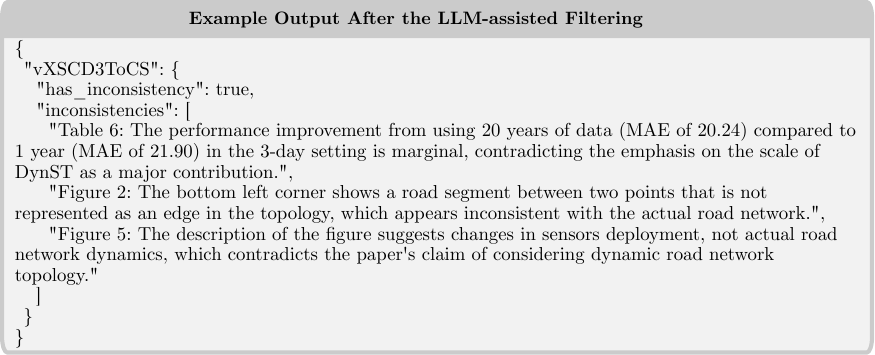}
    \caption{Example output after the LLM-assisted filtering.  }
    \label{fig:example_output_after_llm_assisted_filtering}
\end{figure}

%% file: iclr2026/fig_tex/example_annot_output_in_json_format.tex
\begin{figure}[h]
    \centering
    \includegraphics[width=0.7\linewidth]{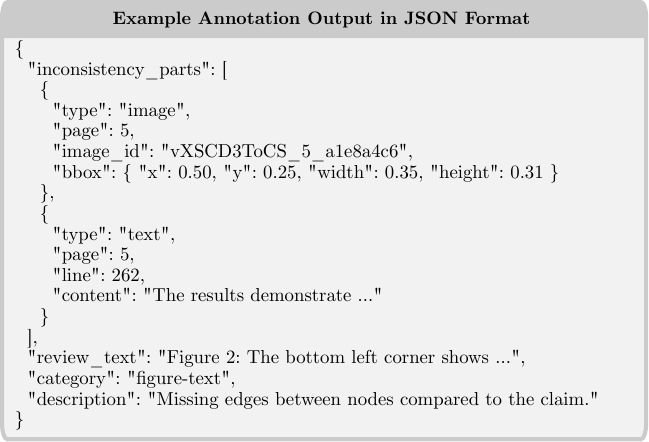}
    \caption{Example annotation output in JSON format.  }
    \label{fig:example_annot_output_in_json_format}
\end{figure}

%% file: iclr2026/fig_tex/distribution_inconsistency_types.tex
\begin{figure}[h]
\begin{center}
\includegraphics[width=0.8\linewidth]{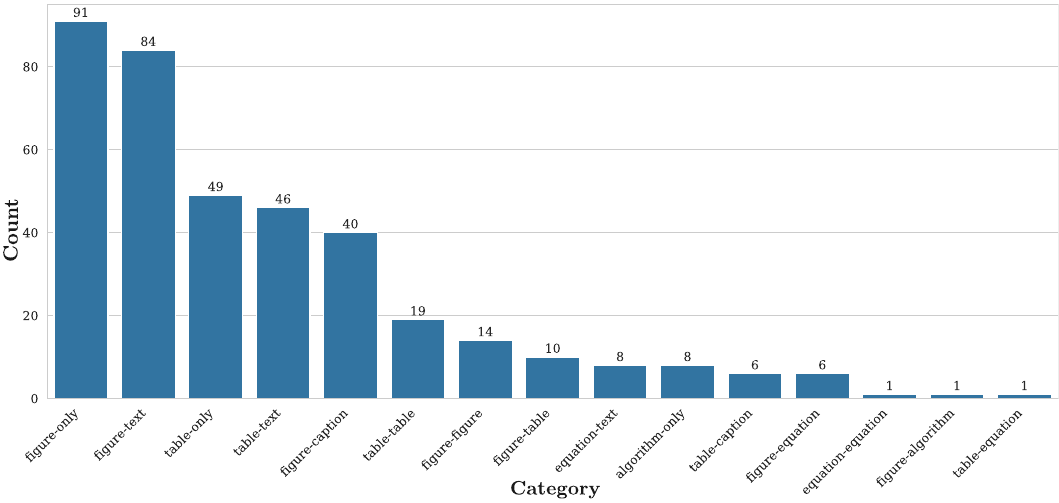}
\end{center}
\caption{Distribution of inconsistency types. We identified 15 categories of inconsistencies based on the elements involved. The most common cases are figure-text mismatches and intra-figure (\textit{figure-only}) inconsistencies.}
\label{fig:type-dist}
\end{figure}

%% file: iclr2026/fig_tex/prompt_preparing_MCQ_inconsistency_identification.tex
\begin{figure}[h]
    \centering
    \includegraphics[width=1\linewidth]{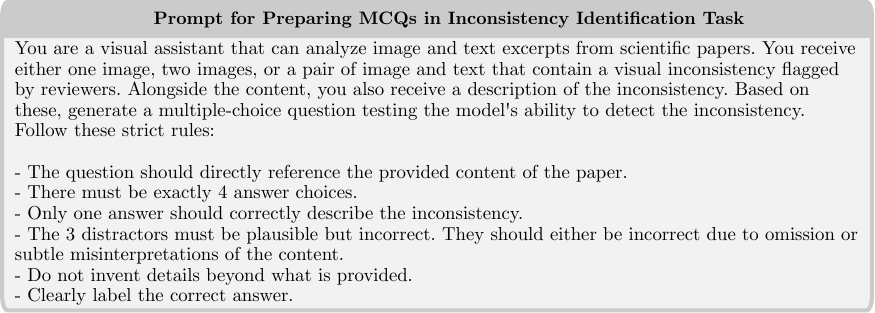}
    \caption{ Prompt for preparing multiple-choice questions in the inconsistency identification task. }
    \label{fig:prompt_preparing_MCQ_inconsistency_identification}
\end{figure}

%% file: iclr2026/fig_tex/example_generated_MCQ_inconsistency_identification.tex
\begin{figure}[h]
    \centering
    \includegraphics[width=1\linewidth]{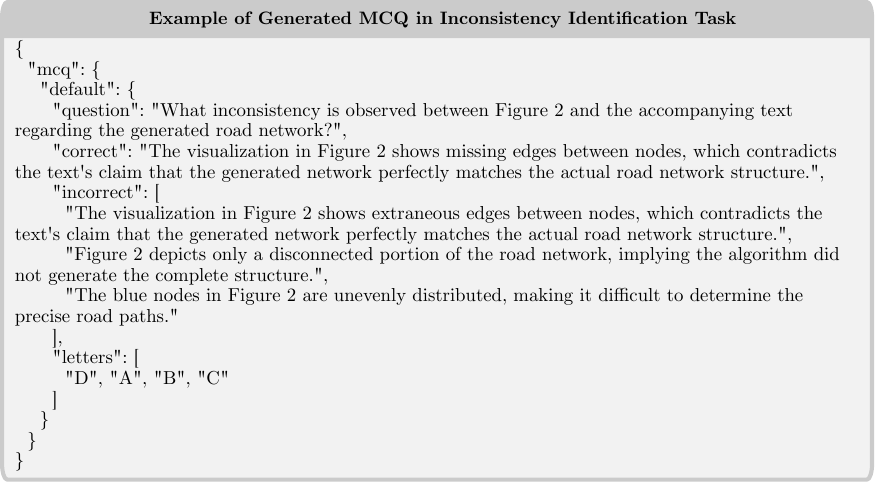}
    \caption{ Example of generated multiple-choice question for the inconsistency identification task. }
    \label{fig:example_generated_MCQ_inconsistency_identification}
\end{figure}

%% file: iclr2026/fig_tex/example_debiased_output_in_evidence_claim_json_format.tex
\begin{figure}[h]
    \centering
    \includegraphics[width=1\linewidth]{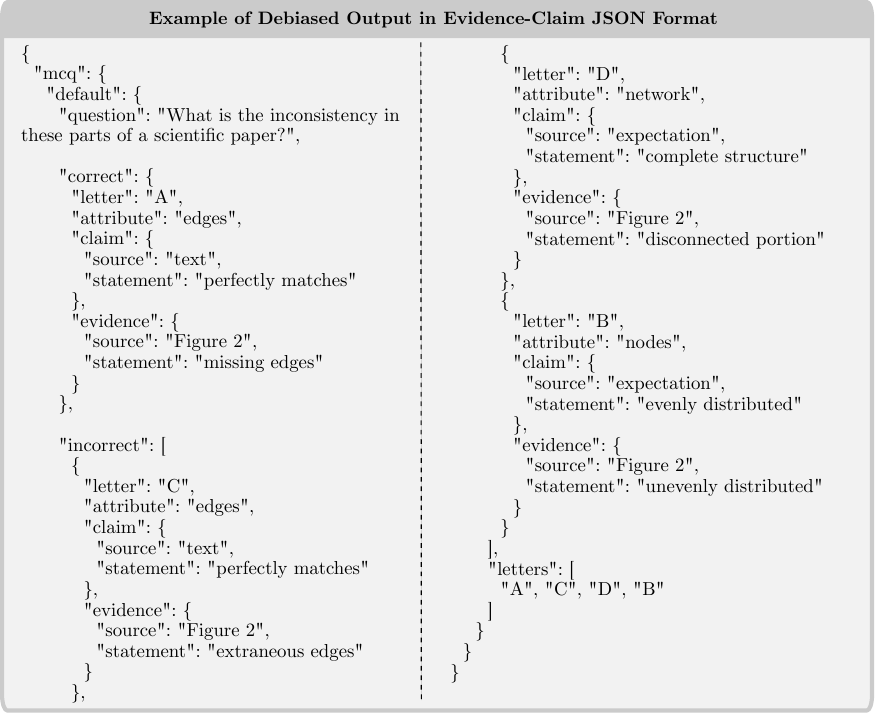}
    \caption{Example of debiased output in evidence-claim JSON format for the inconsistency identification task.  }
    \label{fig:example_debiased_output_in_evidence_claim_json_format}
\end{figure}

%% file: iclr2026/fig_tex/example_output_in_target_action_json_format.tex
\begin{figure}[h]
    \centering
    \includegraphics[width=1\linewidth]{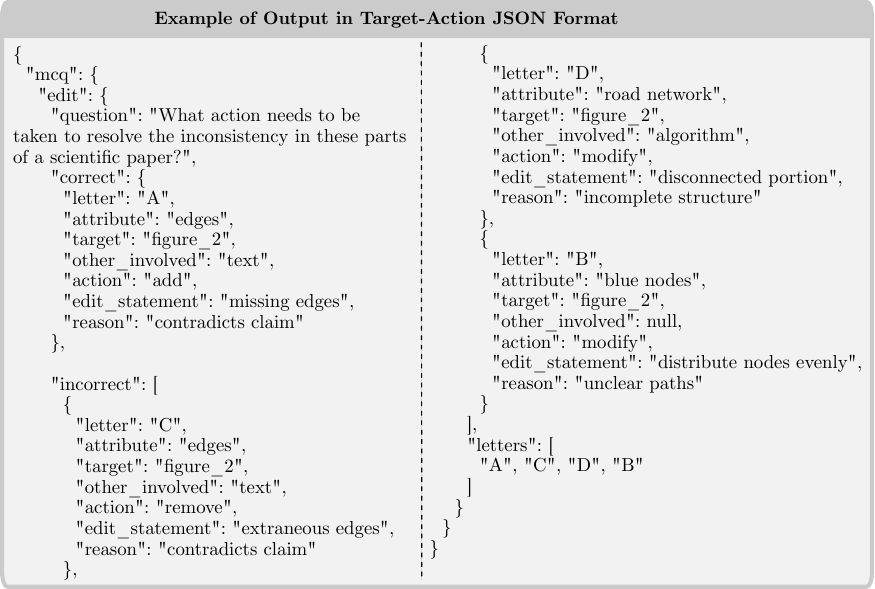}
    \caption{Example of output in target-action JSON format for the inconsistency remedy task, directly converted from the natural language MCQs from the inconsistency identification task. }
    \label{fig:example_output_target_action_json_format}
\end{figure}

%% file: iclr2026/fig_tex/example_output_for_inconsistency_pair_match_task.tex
\begin{figure}[h]
    \centering
    \includegraphics[width=1\linewidth]{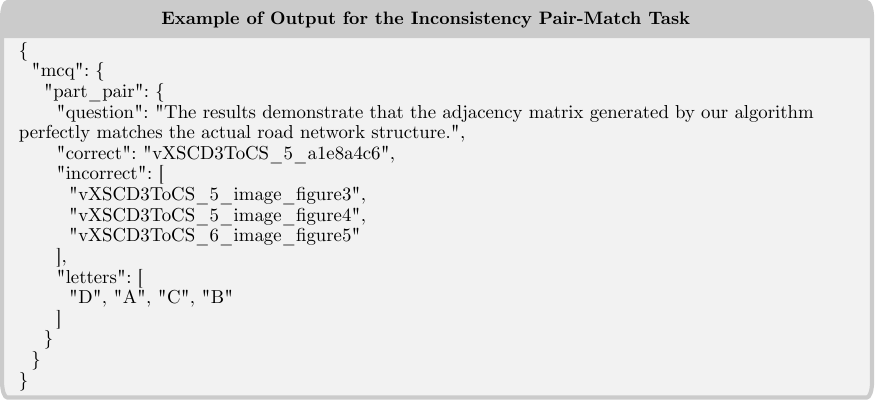}
    \caption{Example output for the inconsistency pair-match task. }
    \label{fig:example_output_for_inconsistency_pair_match_task}
\end{figure}

%% file: iclr2026/tables/user-study-proof.tex
\begin{table}[ht]
\centering
\caption{Model Performance Comparison: User Study vs. Simulation}
\label{tab:user-study-proof}
\small
\resizebox{1\linewidth}{!}{
\begin{tabular}{@{}lcccc@{}}
\toprule
\textbf{\shortstack{Model \\ (Context)}} & \textbf{\shortstack{User Study Subset \\ (Reported)}} & \textbf{\shortstack{Simulation \\ Mean}} & \textbf{\shortstack{Simulation Q3 \\ (75th Percentile)}} & \textbf{\shortstack{Human Perf. \\ (Reference)}} \\
\midrule
\textbf{InternVL 3.5 38B (Focused)} & 71.1\% & 58.6\% & 65.0\% & 77.5\% \\
\textbf{InternVL 3.5 38B (Whole Document)} & 40.5\% & 30.6\% & 35.0\% & 65.0\% \\
\textbf{Qwen 2.5 VL 72B (Focused)} & 65.8\% & 50.5\% & 55.0\% & 77.5\% \\
\textbf{Qwen 2.5 VL 72B (Whole Document)} & 48.6\% & 35.9\% & 40.0\% & 65.0\% \\
\bottomrule
\end{tabular}}
\end{table}

%% file: iclr2026/fig_tex/prompt_llm_based_review_filtering.tex
\begin{figure}[h]
    \centering
    \includegraphics[width=1\linewidth]{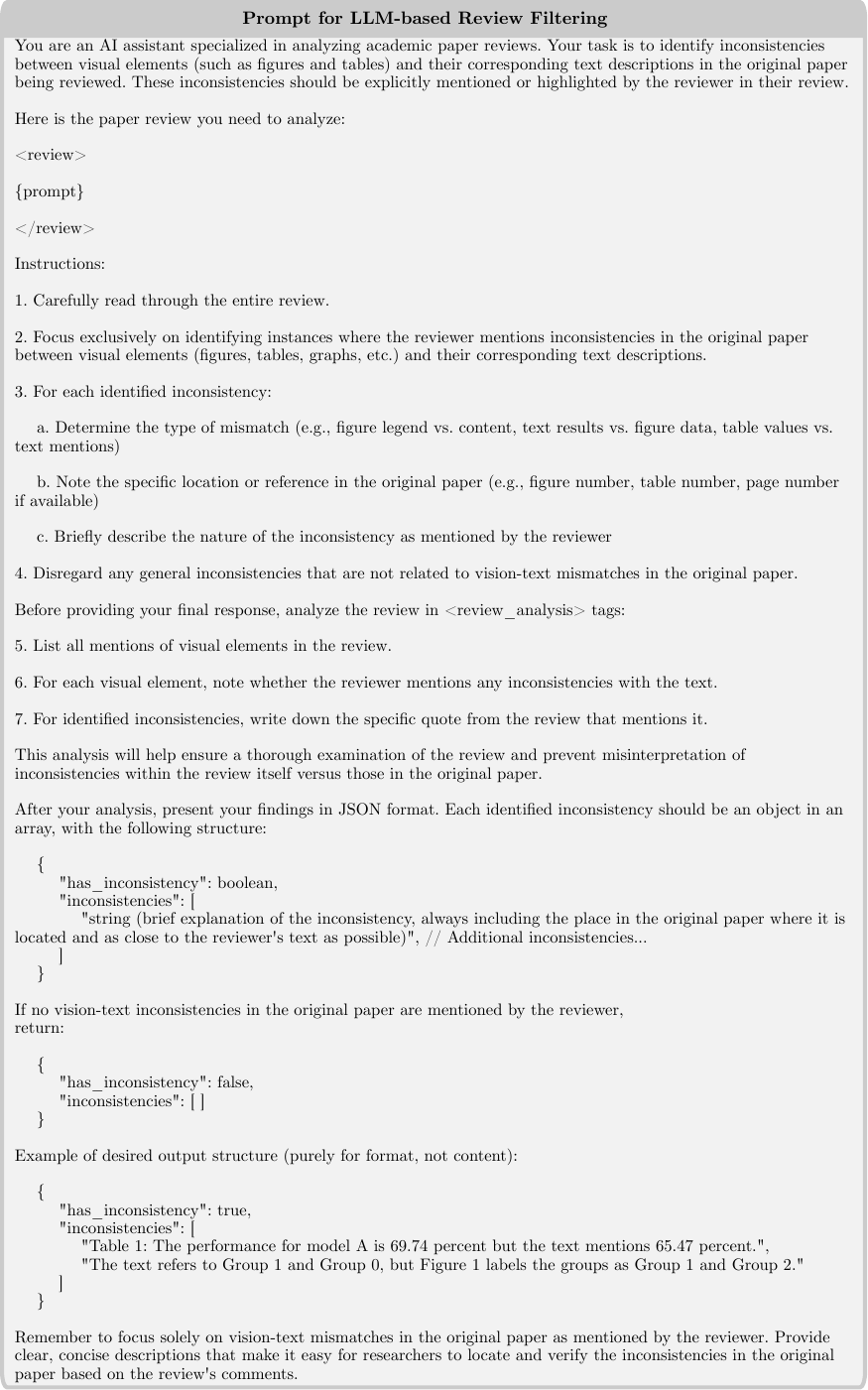}
    \caption{ Prompt for LLM-based review filtering. }
    \label{fig:prompt_llm_based_review_filtering}
\end{figure}

%% file: iclr2026/fig_tex/prompt_conversion_into_evidence_claim_json_format.tex
\begin{figure}[h]
    \centering
    \includegraphics[width=1\linewidth]{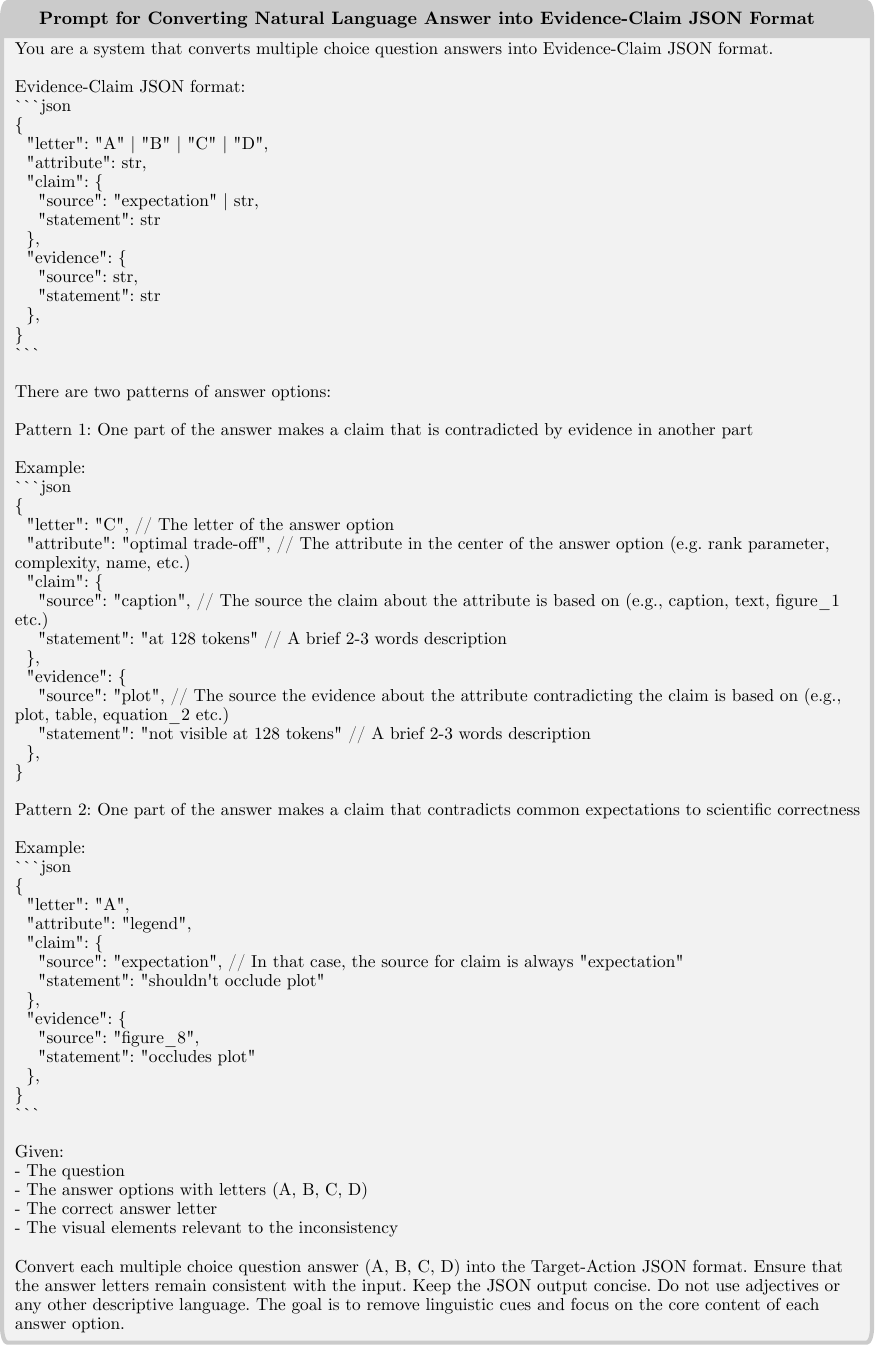}
    \caption{ Prompt for LLM-assisted conversion of natural language answers of the inconsistency identification task into evidence-claim JSON format. The evidence-claim JSON format is used as answer options in the inconsistency identification task. }
    \label{fig:prompt_converting_into_evidence_json_format}
\end{figure}

%% file: iclr2026/fig_tex/prompt_conversion_into_target_action_json_format.tex
\begin{figure}[h]
    \centering
    \includegraphics[width=1\linewidth]{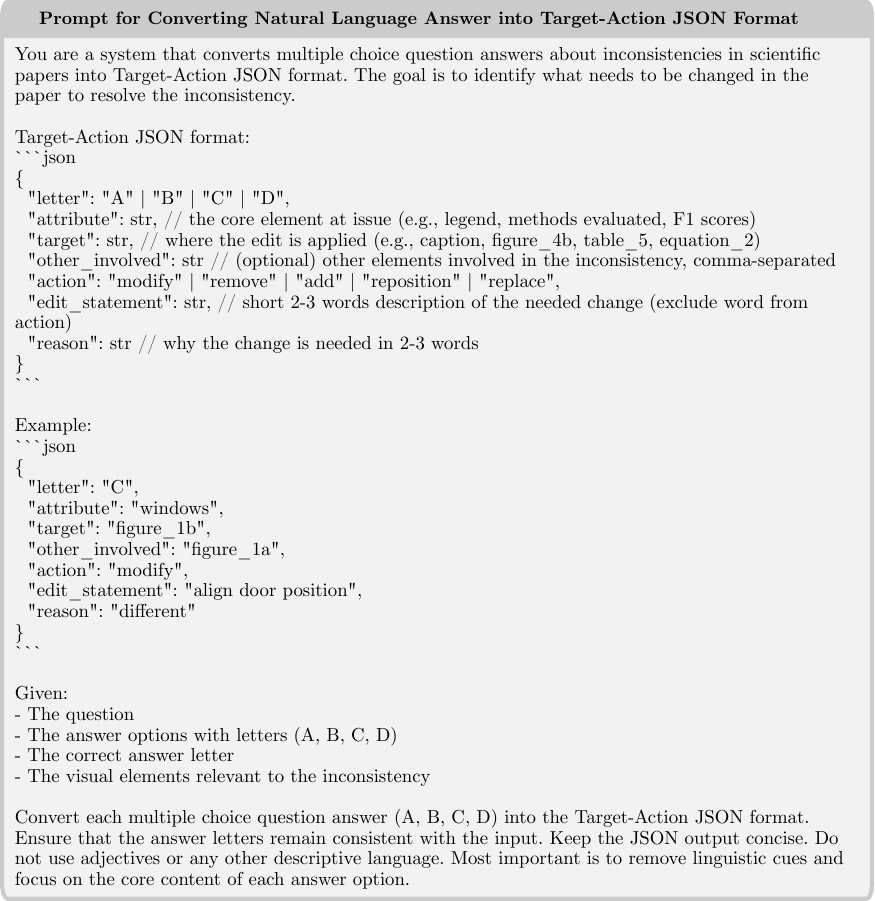}
    \caption{ Prompt for LLM-assisted conversion of natural language answers of the inconsistency identification task into target-action JSON format. The target-action JSON is used as answer options in the inconsistency remedy task. }
    \label{fig:prompt_converting_into_target_action_json_format}
\end{figure}

%% file: iclr2026/fig_tex/annotation_app_vis_first_part.tex
\begin{figure}[h]
    \centering
    \includegraphics[width=1\linewidth]{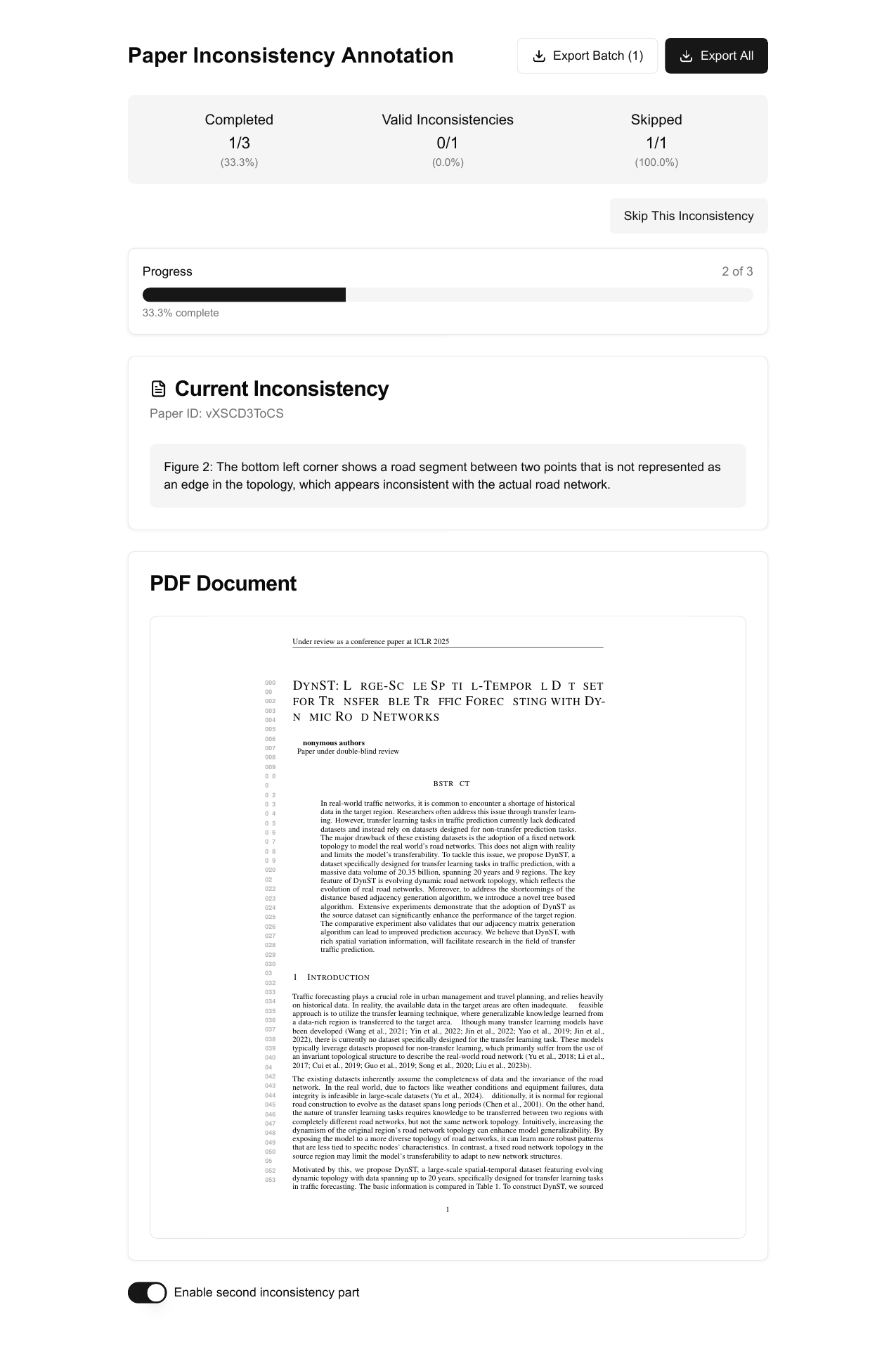}
    \caption{First part of annotation app showing an overview over the annotation progress and embedded original PDF file.}
    \label{fig:ann-app-0}
\end{figure}

%% file: iclr2026/fig_tex/annotation_app_vis_second_part.tex
\begin{figure}[h]
    \centering
    \includegraphics[width=1\linewidth]{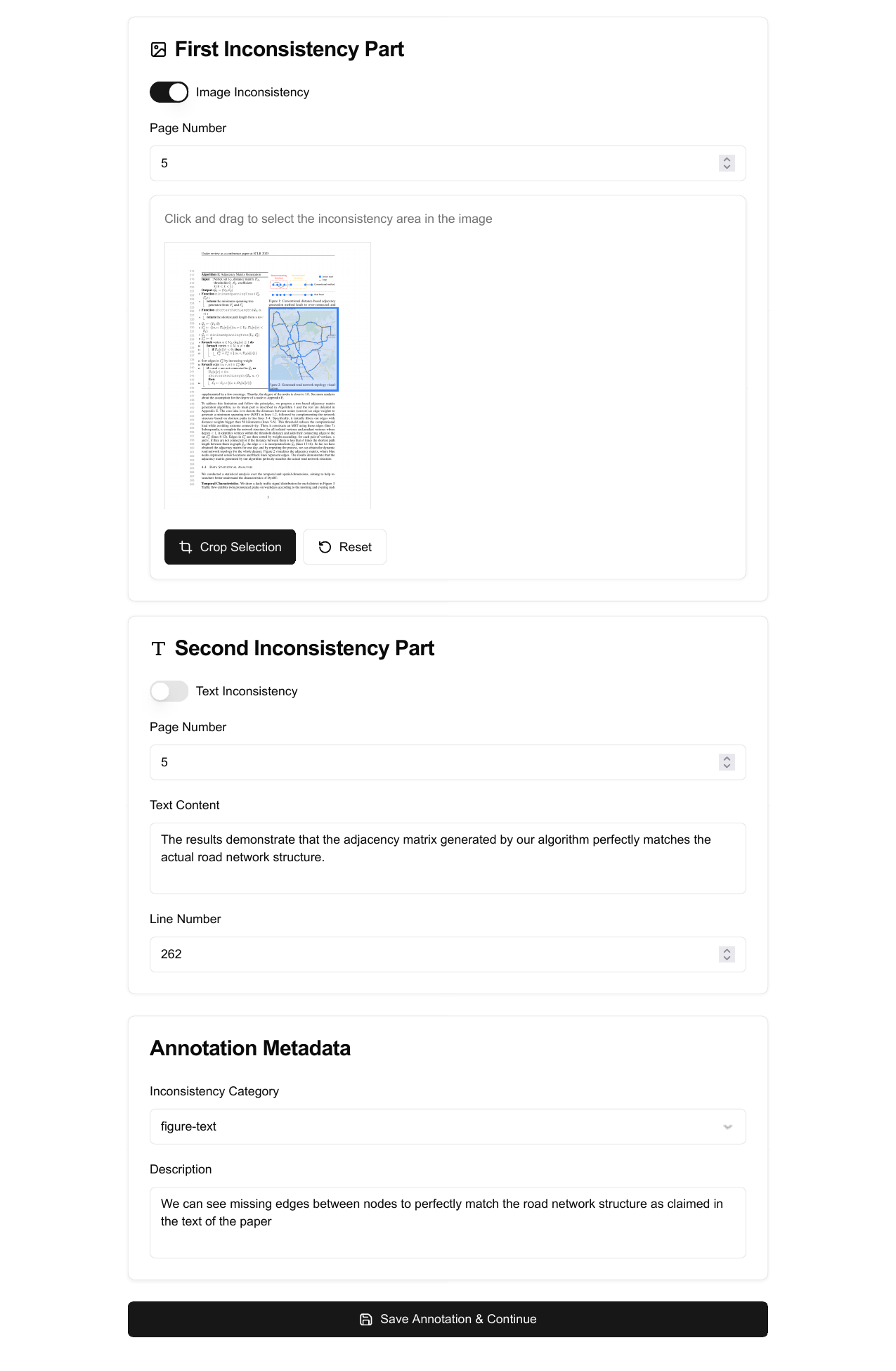}
    \caption{Second part of annotation app for drawing bounding boxes, entering text and further details about the inconsistency.}
    \label{fig:ann-app-1}
\end{figure}

%% file: iclr2026/fig_tex/survey_app_vis_first_part.tex
\begin{figure}[h]
    \centering
    \includegraphics[width=1\linewidth]{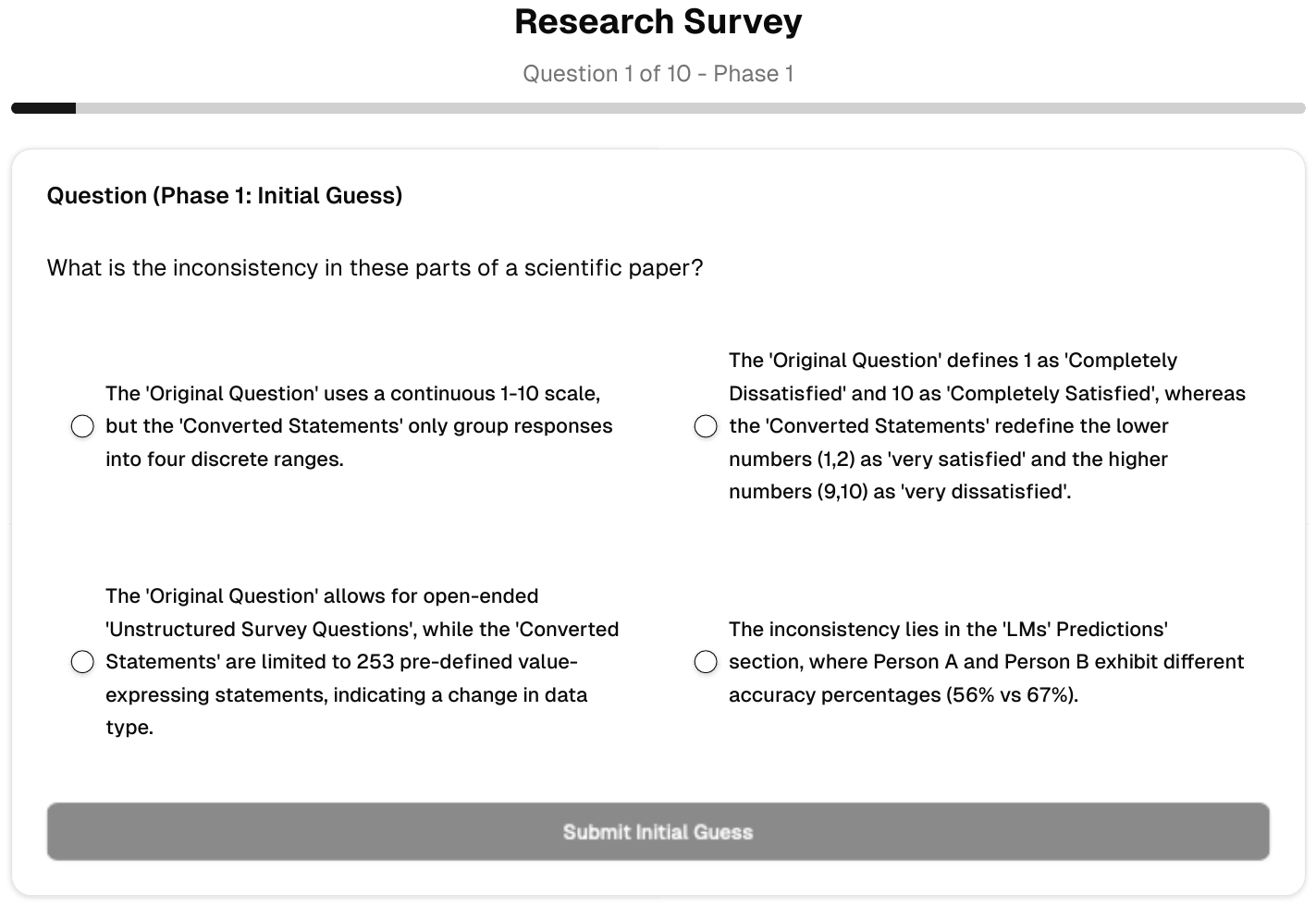}
    \caption{First part of survey interface showing a question with no context provided.}
    \label{fig:survey-app-0}
\end{figure}

%% file: iclr2026/fig_tex/survey_app_vis_second_part.tex
\begin{figure}[h]
    \centering
    \includegraphics[width=1\linewidth]{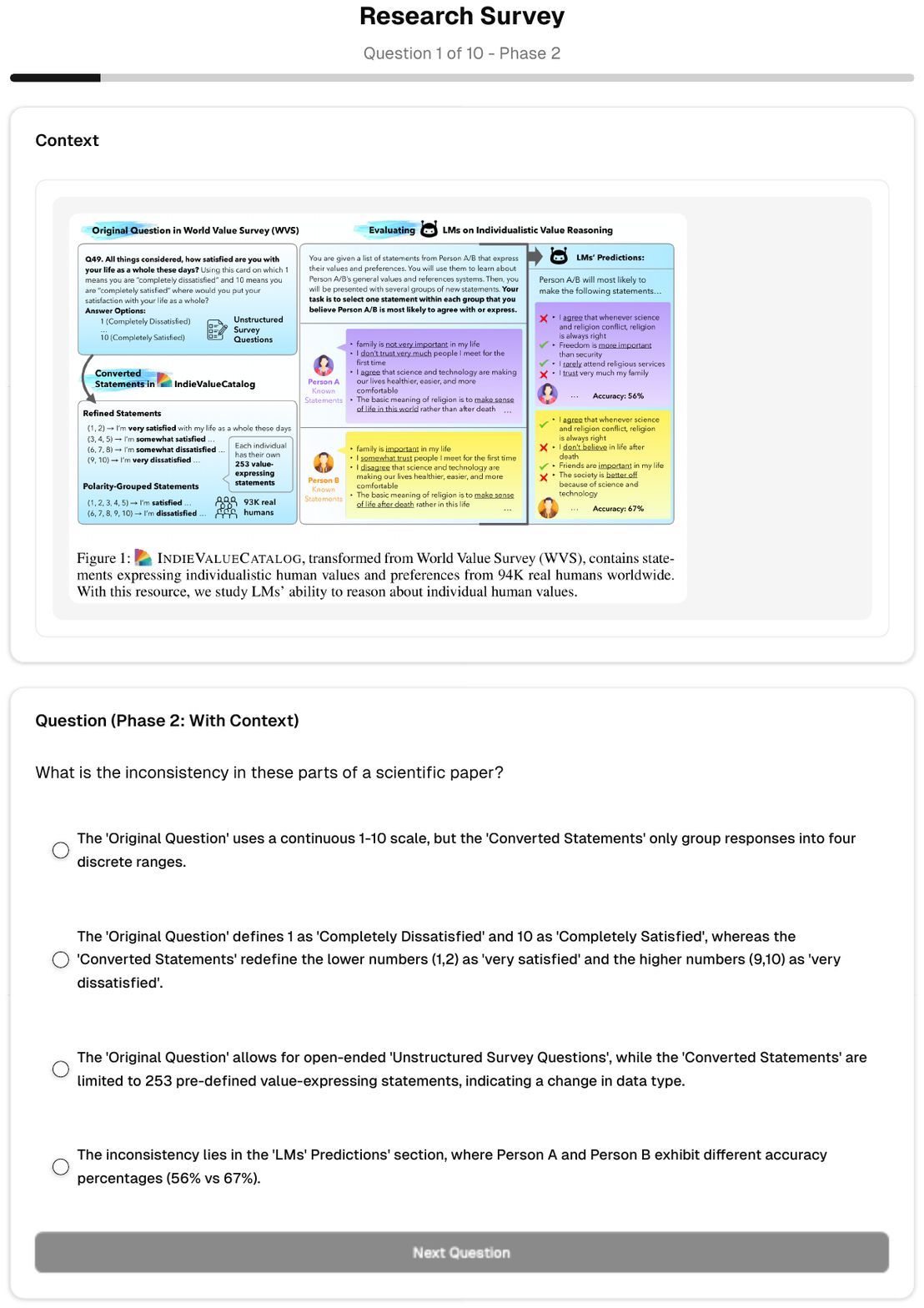}
    \caption{Second part of survey interface showing question with \textit{Focused Context}.}
    \label{fig:survey-app-1}
\end{figure}

%% file: iclr2026/fig_tex/survey_app_vis_third_part.tex
\begin{figure}[h]
    \centering
    \includegraphics[width=1\linewidth]{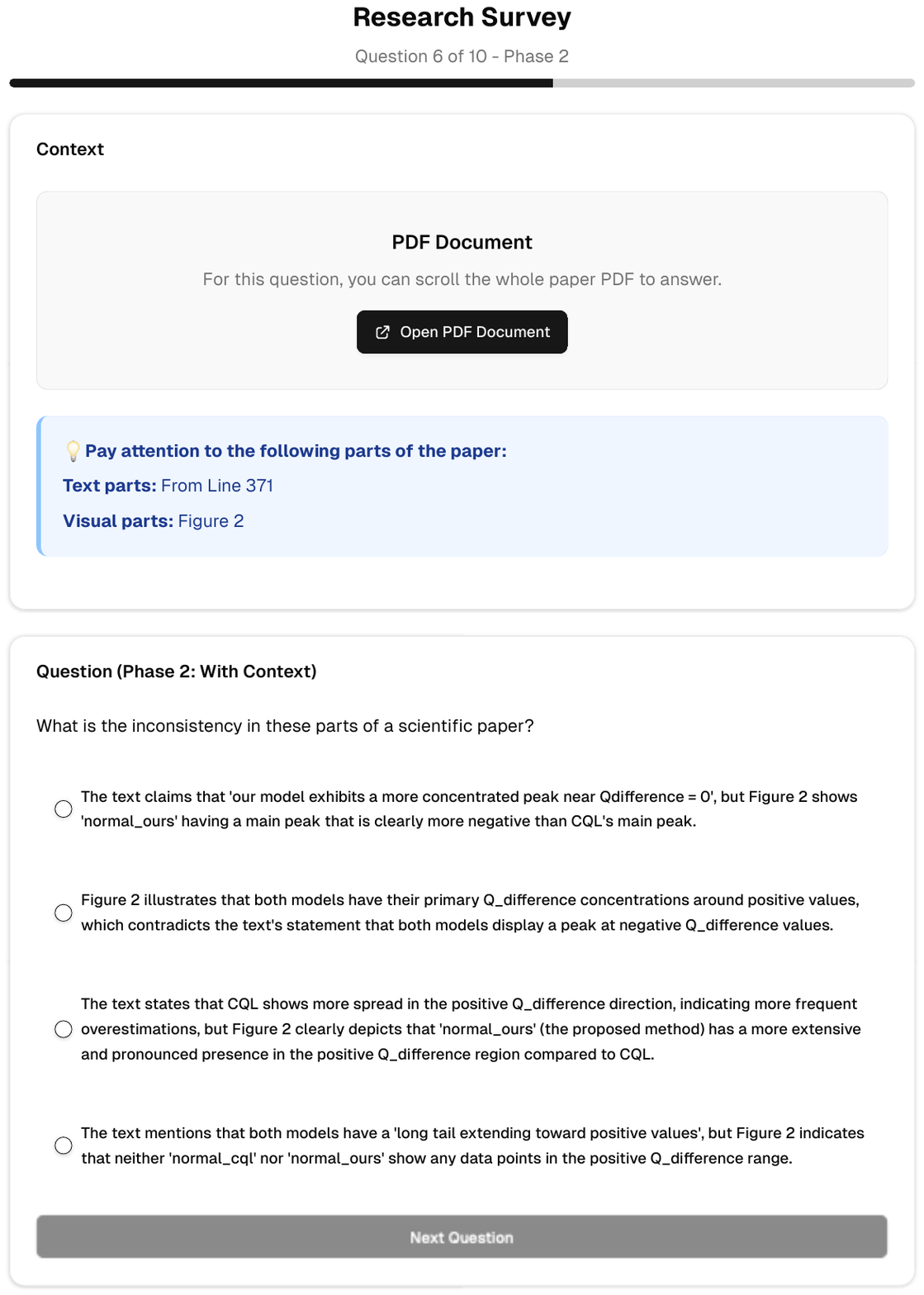}
    \caption{Third part of survey interface showing question with \textit{Full Document Context}.}
    \label{fig:survey-app-2}
\end{figure}